\documentclass[10pt,twocolumn,letterpaper]{article}

\usepackage{cvpr}              

\usepackage{graphicx}
\usepackage{amsmath}
\usepackage{amssymb}
\usepackage{booktabs}

\usepackage{multirow}
\usepackage{makecell}

\usepackage{microtype}

\usepackage[accsupp]{axessibility}  

\usepackage[pagebackref,breaklinks,colorlinks]{hyperref}

\hypersetup{
pdftitle={Non-Probability Sampling Network for Stochastic Human Trajectory Prediction},
pdfsubject={IEEE Conference on Computer Vision and Pattern Recognition},
pdfauthor={Inhwan Bae; Jin-Hwi Park; Hae-Gon Jeon},
}

\usepackage[capitalize]{cleveref}
\crefname{section}{Sec.}{Secs.}
\Crefname{section}{Section}{Sections}
\Crefname{table}{Table}{Tables}
\crefname{table}{Tab.}{Tabs.}

\newcommand{\darrow}{\raisebox{0.15ex}{$\downarrow$}}
\newcommand{\uarrow}{\raisebox{0.15ex}{$\uparrow$}}
\newcommand{\tbf}[1]{\textbf{#1}}
\newcommand{\tul}[1]{\underline{#1}}

\def\cvprPaperID{8037} 
\def\confName{CVPR}
\def\confYear{2022}

\begin{document}

\title{Non-Probability Sampling Network for Stochastic Human Trajectory Prediction}

\author{Inhwan Bae, Jin-Hwi Park and Hae-Gon Jeon\thanks{Corresponding author}\\
AI Graduate School, GIST, South Korea\\
{\tt\small \{inhwanbae, jinhwipark\}@gm.gist.ac.kr, haegonj@gist.ac.kr} 
}
\maketitle

\begin{abstract}
Capturing multimodal natures is essential for stochastic pedestrian trajectory prediction, to infer a finite set of future trajectories. The inferred trajectories are based on observation paths and the latent vectors of potential decisions of pedestrians in the inference step. However, stochastic approaches provide varying results for the same data and parameter settings, due to the random sampling of the latent vector. In this paper, we analyze the problem by reconstructing and comparing probabilistic distributions from prediction samples and socially-acceptable paths, respectively. Through this analysis, we observe that the inferences of all stochastic models are biased toward the random sampling, and fail to generate a set of realistic paths from finite samples. The problem cannot be resolved unless an infinite number of samples is available, which is infeasible in practice. We introduce that the Quasi-Monte Carlo (QMC) method, ensuring uniform coverage on the sampling space, as an alternative to the conventional random sampling. With the same finite number of samples, the QMC improves all the multimodal prediction results. We take an additional step ahead by incorporating a learnable sampling network into the existing networks for trajectory prediction. For this purpose, we propose the Non-Probability Sampling Network (NPSN), a very small network ($\sim$5K parameters) that generates purposive sample sequences using the past paths of pedestrians and their social interactions. Extensive experiments confirm that NPSN can significantly improve both the prediction accuracy (up to 60\%) and reliability of the public pedestrian trajectory prediction benchmark. Code is publicly available at \url{https://github.com/inhwanbae/NPSN}.
\end{abstract}


\vspace{-1mm}
\section{Introduction}
The goal of predicting pedestrian trajectories is to infer socially-acceptable paths based on previous steps while considering the social norms of other moving agents. Many earlier works~\cite{helbing1995social, 5459260, 5206641, yamaguchi2011you} on human trajectory prediction are based on deterministic approaches which yield the most likely single path. One of the earliest works in~\cite{helbing1995social} models a social force using attractive and repulsive forces between pedestrians. Since then, motion time-series and agent interactions have been applied to trajectory forecasting. With the development of recurrent neural networks (RNNs), pioneering works such as, Social-LSTM~\cite{alahi2016social} and Social-Attention~\cite{vemula2018social}, have adopted a social pooling and attention mechanisms between spatial neighbors. These approaches have become baseline models in areas such as spatial relation aggregation~\cite{gupta2018social, huang2019stgat, Shi2021sgcn, salzmann2020trajectron++, mohamed2020social, sun2020rsbg} and temporal future prediction~\cite{mangalam2020pecnet, sun2020reciprocal, zhao2020tnt, Lee_2017_CVPR, Marchetti_2020_CVPR, zhang2019sr}.

\begin{figure}[t]
    \vspace{-0.5mm}
    \centering
    \includegraphics[width=1\linewidth,trim={25mm 0 25mm 0},clip]{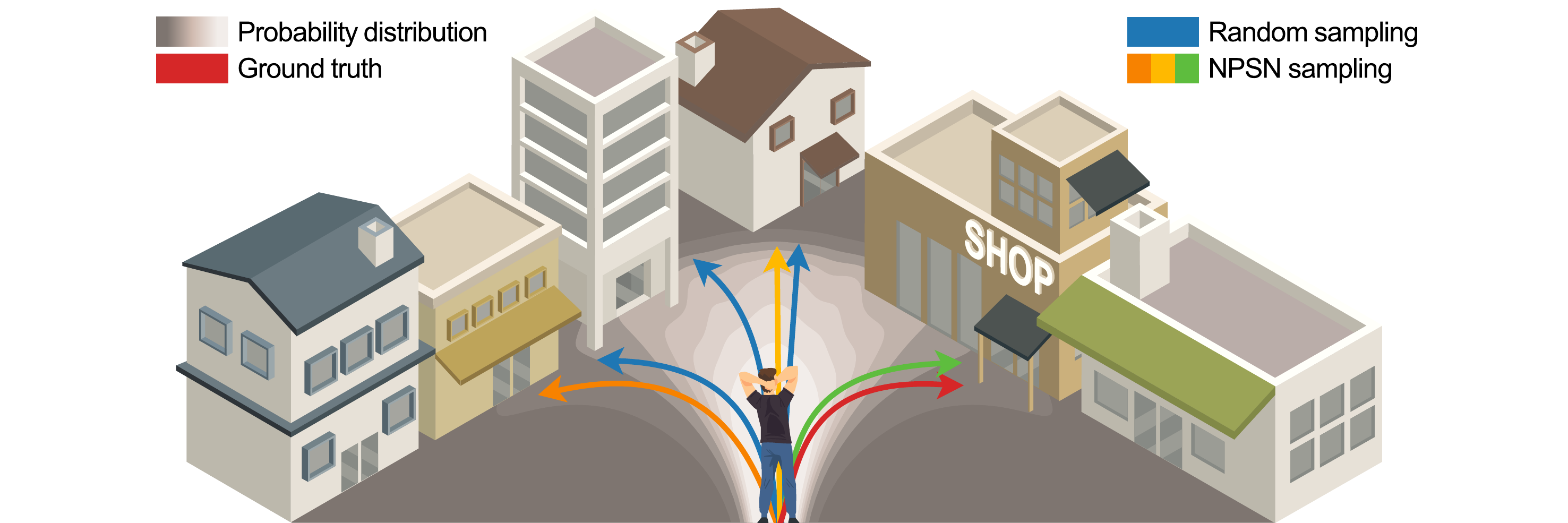}
    \vspace{-6.5mm}
    \caption{An illustration of a probability distribution of stochastic trajectory prediction and selected paths from each sampling method. While the trajectories from the random sampling are biased in that they do not consider space of all possible distributions, our NPSN purposively generates the accurate route, turning to SHOP, even with its low probability.}
    \vspace{-2.5mm}
    \label{fig:teaser}
\end{figure}

Recently, generative models, which infer the distribution of potential future trajectories, are likely to inspire a major paradigm shift away from the single best prediction methods~\cite{gupta2018social,liang2019peeking,li2019conditional,shi2020multimodal,sadeghian2019sophie,kosaraju2019social,sun2020reciprocal,dendorfer2021mggan,zhao2019matf,tao2020dynamic,sun2020rsbg,shafiee2021Introvert,Lee_2017_CVPR,Ivanovic_2019_ICCV,salzmann2020trajectron++,huang2019stgat,mohamed2020social,liang2020garden,Shi2021sgcn,yu2020spatio,li2020Evolvegraph,mangalam2020pecnet,liu2021causal,liu2020snce}. The generative models represent all possible paths, such that pedestrians may go straight, turn left/right at an intersection or take a roundabout way to avoid obstacles. To efficiently establish this multi-modality, a stochastic process is introduced to the trajectory prediction~\cite{gupta2018social}, which models the inferred uncertainty of pedestrians' movements in every time frame. Stochastic trajectory prediction models start by generating a random hypothesis. Due to the non-deterministic nature of random sampling, the quality of the hypotheses depends on the number of samples. Ideally, an infinite number of hypotheses would be able to characterize all possible movements of pedestrians, but this is infeasible. In practice, a fixed number of multiple trajectories are randomly sampled using the Monte Carlo (MC) method, and all existing stochastic models follow this random sampling strategy. However, the number of samples is typically too small to represent socially-acceptable pedestrian trajectories because they are biased toward the random sampling, as illustrated in~\cref{fig:teaser}.

\begin{figure}[t]
\vspace{-1mm}
\centering
\includegraphics[width=1\linewidth]{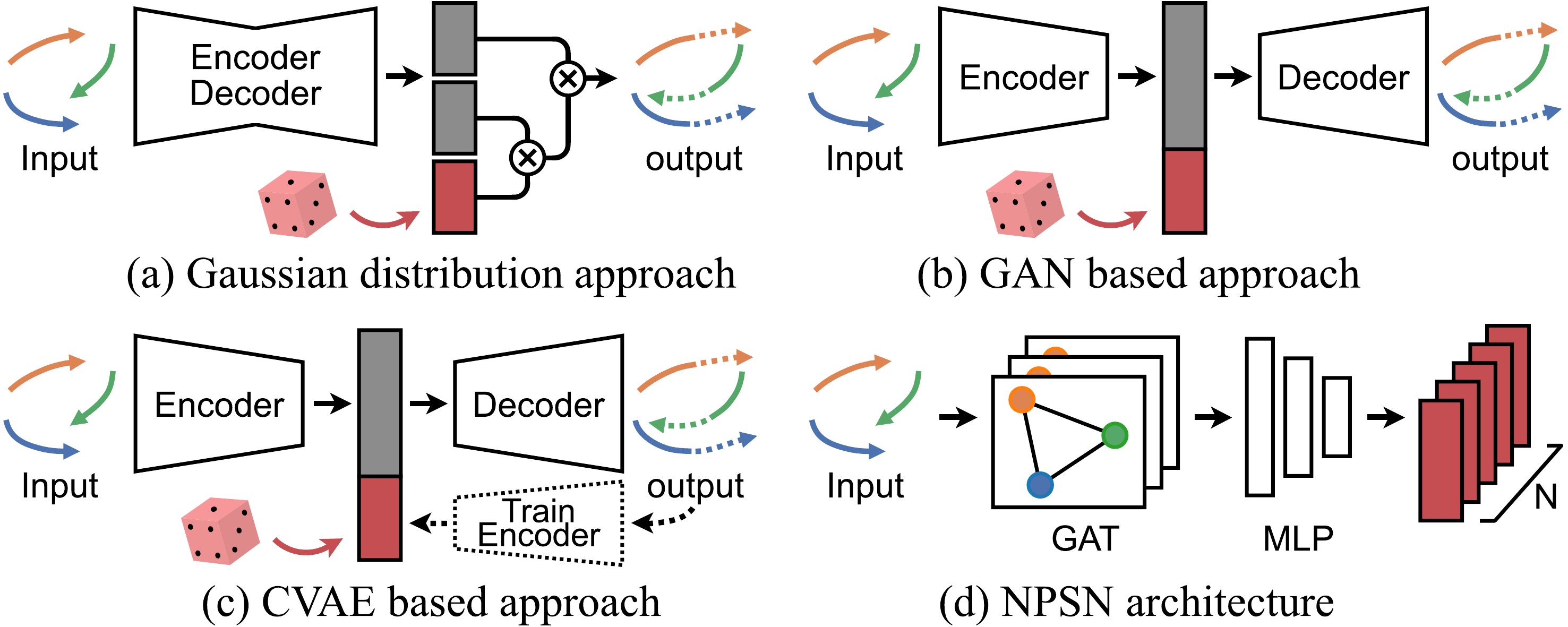}
\vspace{-7mm}
\caption{Illustrations of stochastic human trajectory prediction and our NPSN method. The red box indicates the latent vector.}
\vspace{-2mm}
\label{fig:noise_models}
\end{figure}

In this paper, we revisit the state-of-the-art works which employ the stochastic process for multimodal prediction (\cref{fig:noise_models}-(a)$\sim$(c)). We prove that all of the expected values in the generated trajectory distributions with Generative Adversarial Networks (GANs)~\cite{gupta2018social, huang2019stgat, liu2021causal}, Conditional Variational Auto-Encoders (CVAEs)~\cite{salzmann2020trajectron++, mangalam2020pecnet, liu2020snce}, and Gaussian methods~\cite{mohamed2020social, Shi2021sgcn} are biased. Afterward, we introduce a Quasi-Monte Carlo (QMC) sampling method that effectively alleviates this problem using a low-discrepancy sequence, instead of random sampling. Lastly, we push the random sampling forward with a learnable method: Non-Probability Sampling Network (NPSN), a very small network that generates purposive sample sequences using observations and agent interactions in \cref{fig:noise_models}-(d). Without structurally modifying the existing models in any way, we achieve significant improvements in the performance of pedestrian trajectory prediction. 
This is accomplished by replacing one line of code on random sampling with our NPSN. Interestingly, one of the existing models using our NPSN as an auxiliary module achieves the best performance in all evaluation metrics.

Unlike previous methods, the proposed approach focuses on the sampling method to generate a set of random latent vectors. To the best of our knowledge, our work is the first attempt to adopt QMC sampling and to propose a learnable method for purposive sampling in trajectory forecasting in~\cref{fig:teaser}.

\vspace{-0.3mm}
\section{Related Works}
\vspace{-0.3mm}
\subsection{Stochastic trajectory prediction}
\vspace{-0.3mm}
Convolutional neural network (CNN)-based approaches using Gaussian distribution have improved the efficiency of pedestrian trajectory prediction. Social-LSTM~\cite{alahi2016social}, a pioneering model in this field, predicts a bivariate Gaussian distribution consisting of five parameters for the observed trajectories of pedestrians. However, it has a limitation when inferring single paths, since it only selects the best one sample from the distribution in inference time. 
Follow-up works~\cite{vemula2018social,mohamed2020social,Shi2021sgcn,shi2020multimodal} predict multiple paths by sampling multiple next coordinates based on predicted distributions.

As another methodology, a generative model is introduced to predict realistic future paths. Social-GAN~\cite{gupta2018social} firstly uses a generative framework that recursively infers future trajectory. The benefit of GAN is that it generates various outputs according to latent vectors. As a result, inter-personal, socially acceptable and multimodal human behaviors are accounted for in the pedestrian trajectory prediction. Such a research stream encourages to define a variety loss which calculates for the best prediction among multiple samples for diverse sample generation.~\cite{kosaraju2019social, sadeghian2019sophie, sun2020reciprocal, liu2021causal, dendorfer2021mggan, huang2019stgat}.

Similarly, there have been attempts to predict diverse future generations using CVAE frameworks.
DESIRE~\cite{Lee_2017_CVPR} uses a latent variable to account for the ambiguity of future paths and learns a sampling model to produce multiple hypotheses of future trajectories from given observations. This approach provides a diverse set of plausible predictions without the variety loss, and shares inspiration to objectives in many CVAE-based models~\cite{salzmann2020trajectron++, mangalam2020pecnet, yu2020spatio,liu2020snce,Ivanovic_2019_ICCV}.

All of these methods include a random sampling process and are sensitive to bias, due to the fixed number of samples, as above mentioned. In addition, current state-of-the-art models with CVAE frameworks outperform Gaussian distribution-based methods~\cite{mohamed2020social, Shi2021sgcn}. In this study, we analyze these phenomena with respect to the bias of stochastic trajectory prediction, and show that the Gaussian distribution-based approaches achieve noticeable performance improvements by minimizing the bias, even better than the CVAE-based methods. Lastly, we mention a recent deterministic approach~\cite{zhao2020tnt} that predicts multiple trajectories, which is beyond the scope of this paper.

\vspace{-0.7mm}
\subsection{Learning latent variables}
\vspace{-0.7mm}
Some works account for the transformation of latent spaces by using prior trajectory information. PECNet~\cite{mangalam2020pecnet} for example uses a truncation trick in latent space to adjust the trade-off between the fidelity and the variety of samples. In their learning approach, both IDL~\cite{li2019idl} and Trajectron++~\cite{salzmann2020trajectron++} predict the mean and standard deviation of a latent distribution in an inference step. Rather than directly predicting the distribution parameters, AgentFormer~\cite{yuan2021agent} uses a linear transform of Gaussian noise to produce the latent vector. These methodologies still run the risk of bias because of the random sampling of the latent vectors. In the present work, we aim to reduce the bias using a discrepancy loss of a set of sampled latent vectors.

\vspace{-0.7mm}
\subsection{Graph-based approaches}
\vspace{-0.7mm}
Pioneering works have introduced the concepts of social-pooling~\cite{alahi2016social, gupta2018social, sun2020reciprocal} and social-attention mechanisms~\cite{vemula2018social, zhang2019sr, li2020Evolvegraph} to capture the social interactions among pedestrians in scenes. Recently, Graph Neural Network (GNN)-based approaches~\cite{huang2019stgat, kosaraju2019social, mohamed2020social, liang2020garden, Shi2021sgcn, li2020Evolvegraph, Bae_Jeon_2021} have been introduced to model agent-agent interactions with graph-based policies. In the GNN-based works, pedestrians are regarded as nodes of the graph, and their social relations are represented as edge weights. Social-STGCNN~\cite{mohamed2020social} presents a Graph Convolutional Network (GCN)~\cite{kipf2016semi}-based trajectory prediction which aggregates the spatial information of distances among pedestrians. Graph Attention Networks (GATs)~\cite{velivckovic2018graph} implicitly assign more weighting to edges with high social affinity on the pedestrian graph~\cite{huang2019stgat,kosaraju2019social,sun2020rsbg,Shi2021sgcn,yu2020spatio}. Multiverse~\cite{liang2020garden} and SimAug~\cite{liang2020simaug} utilize GATs on 2D grids to infer feasible trajectories. Unlike these previous works, where GATs are used in the encoding process, we apply a GAT framework to a sampling process on the latent space to make a decoder predict future paths more accurately.

\vspace{-0.5mm}
\subsection {Monte Carlo Sampling Method}
\vspace{-0.5mm}
(Quasi-) Monte Carlo is a computational technique for numerical experiment using random numbers. Exploiting the random numbers allows one to approximate integrals, but this is highly error prone. The error directly depends on the random sampling methods from probability distributions. QMC sampling is developed with quasi-random sequences, known as low-discrepancy sequences~\cite{low_discrepancy} and is generated in a deterministic manner. It is widely utilized for many computer vision tasks, such as depth completion~\cite{DC_MC_2}, 3D reconstruction~\cite{MC_3Drecon,MC_3DPointCloudRegist}, motion tracking~\cite{MC_motionTracking} and neural architecture search~\cite{MC_NAS_1,MC_NAS_2}. We firstly apply QMC sampling to ensure uniform coverage of the sampling spaces for pedestrian trajectory prediction. Note that the sequence is uniformly distributed if the discrepancy tends to be zero, as the number of samples goes to infinity.

\begin{figure}[t]
\centering
\vspace{-1mm}
\includegraphics[width=\columnwidth]{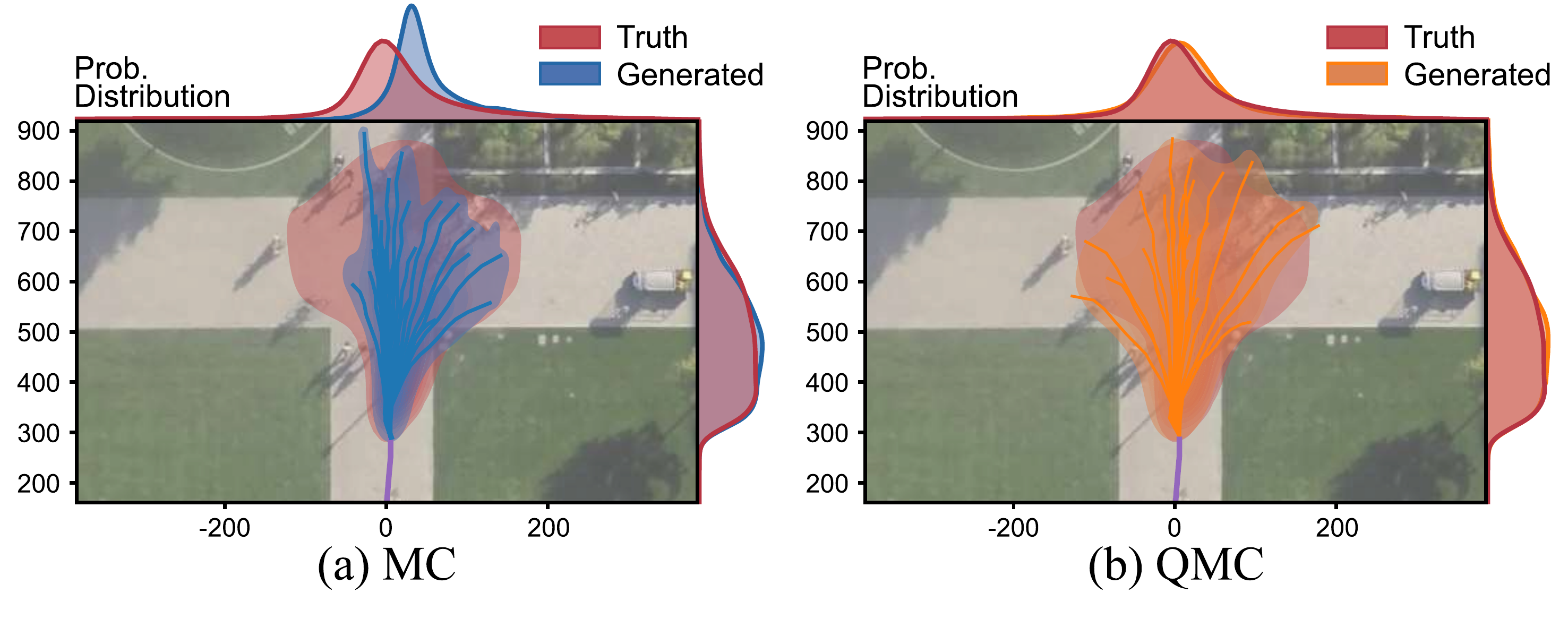}
\vspace{-9mm}
\caption{An example of the probability map of the truth plausible trajectory distribution and the generated distributions with $N\!\!\!\;=\!20$ samples $\smash{\hat{Y}_{l}^{1:T_{pred}}}$ using MC and QMC method on PECNet~\cite{mangalam2020pecnet}.}
\vspace{-2mm}
\label{fig:2d_distribution}
\end{figure}

\vspace{-0.5mm}
\section{Generated Trajectories Are Biased}
\vspace{-0.5mm}
In this section, we start with the problem definition for pedestrian trajectory prediction in~\cref{subsec:problem}. We then theoretically demonstrate that generated trajectories from stochastic trajectory prediction models are biased toward random sampling in~\cref{subsec:bias}. We also introduce a way to alleviate the bias with a low-discrepancy sequence for stochastic prediction in~\cref{subsec:QMC}.

\subsection{Problem Definition}
\vspace{-0.5mm}
\label{subsec:problem}
\microtypesetup{disable}
We formulate the pedestrian trajectory prediction task as a multi-agent future trajectory generation problem conditioned on their past trajectories. To be specific, during the observation time frames $1\kern-0.6ex\leq\kern-0.6ex t\kern-0.6ex\leq\kern-0.6ex T_{obs}$, there are $L$ pedestrians in a scene. The observed trajectory sequence is represented as $\smash{X_{l}^{1:T_{obs}}}\!\!=\!\{ X_l^t|t \in [1, ..., T_{obs}] \}$ for $\forall l\!\in\![1, ..., L]$, where $x_l^t$ is the spatial coordinate of each pedestrian $l$ at time frame $t$. With the given observed sequence, the goal of the trajectory prediction is to learn potential distributions to generate $N$ plausible future sequences $\smash{\hat{Y}_{l}^{1:T_{pred}}}\!\!=\!\{\hat{Y}_{l, n}^{t} | t \in [1, ..., T_{pred}], n \in [1, ..., N]\}$ for all $L$ pedestrians.
\microtypesetup{enable}

\vspace{-0.5mm}
\subsection{Stochastic Trajectory Prediction is Biased.} 
\vspace{-0.5mm}
\label{subsec:bias}
The generated trajectory $\smash{\hat{Y}_{l}^{1:T_{pred}}}$ comes from a distribution of possible trajectories which are constructed by pedestrians' movements based on social forces~(\cref{fig:2d_distribution}). $\mathcal{T}_{truth}$ is an expectation value computed with a plausible trajectory distribution, and $\mathcal{T}_{gen}$ is calculated with $\smash{\hat{Y}_{l}^{1:T_{pred}}}$ of $N$ which are independent and identically distributed (IID) random samples, \ie the term is random if one uses different samples to generate trajectories. The expectation $\mathcal{T}_{gen}$ is a Monte Carlo estimate of integral, \ie relevant expectation.

Suppose that the expectation we want to compute from the trajectory distribution is $\smash{I(\tau) = \int_{[0,1]^s} \tau(x)q(x)dx}$ which is the expected value of $\tau(x)$ for random variable $x$ with a density $q$ on $s$-dimensional unit cube $[0,1]^s$. Then, the Monte Carlo estimator for the generated trajectory distribution with $N$ samples can be formulated as below:
\noindent\vspace{-2.5mm}
\begin{equation}
    \hat{I}(\tau) = \hat{I}_{N,s}(\tau) = \frac{1}{N} \sum_{i=1}^{N} \tau(x_i),
    \vspace{-2mm}
\end{equation}
\vspace{-2mm}
\begin{equation}
    Pr(\lim_{n \to \infty} \hat{I}(\tau) = I(\tau)) = 1,    
    \label{eq:probability}
    \vspace{-0.5mm}
\end{equation}
where $Pr(\cdot)$ denotes a probability. 

By the Strong Law of large numbers~\cite{LawofLargeNumber}, the MC estimate converges to $I(\tau)$ as the number of samples $N$ increases without bound. Now, we assume that $\tau(x)$ has a finite variance $K(\tau)$ and define the error $\alpha$ as below:
\noindent\vspace{-1.5mm}
\begin{equation}
    \alpha = \hat{I}_{N,s}(\tau) - I(\tau),    
    \vspace{-1mm}
\end{equation}
\vspace{-5mm}
\begin{equation}
    \mathbb{E}[\alpha] = 0,\quad var(\alpha) = \frac{K(\tau)}{N},    
    \vspace{-0.5mm}
\end{equation}
where $\mathbb{E}$ is an expectation and $K(t)$ is $\int (\tau(x) - I(\tau))^2q(x) dx$. Note that the $K(\tau)$ is non-negative and depends on the function being integrated. The algorithmic goal is to specify the procedure that results in lower variance estimates of the integral. 

Now consider a function of the generator $F$, which is sufficiently smooth, in a Monte Carlo integral $I(\tau)$. We apply the Taylor series expansion of $F(I(\tau)\!+\!\alpha)$ as follows:
\noindent\vspace{-4mm}
\begin{equation}
    F(\hat{I}_{N,s}(\tau)) = F(I(\tau)\!+\!\alpha) \qquad\qquad\qquad\qquad\qquad\quad
\vspace{-2.5mm}
\end{equation}
\begin{equation*}
    \qquad\qquad\quad\; \approx F(I(\tau))\!+\!\alpha F'(I(\tau))\!+\!\alpha^2 \frac{F''(I(\tau))}{2}\!+\!O(\alpha^3).
\vspace{-2mm}
\end{equation*}

\noindent
Therefore, the expectation value of $F(\hat{I}_{N,s}(\tau))$ can be formulated as below:

\noindent\vspace{-2mm}
\begin{equation}
    \mathbb{E}[F(\hat{I}_{N,s}(\tau))] = F(I(\tau)) + \frac{M}{N} + O(\frac{1}{N^2}),
    \vspace{-0.5mm}
\end{equation}
where $M = K(\tau)(F''(I(\tau))/2)$ and the $\frac{M}{N}$ is a bias. Since the term $\mathcal{T}_{gen}$ is estimated with an MC integration, the estimate must have a bias of $\frac{M}{N} + O(\frac{1}{N^2})$. Note that the bias in the generated trajectories vanishes for $N \rightarrow \infty$, however, it is infeasible to utilize all infinite possible paths in practice. Since $M$ depends on the generator, the generated trajectories are differently biased depending on the number of generated samples as well as the generators, which is validated in \cref{subsec:method}.

\subsection{Quasi-Monte Carlo for Trajectory Prediction}
\label{subsec:QMC}

The QMC method utilizes a low discrepancy sequence including the Halton sequence~\cite{halton} and the Sobol sequence~\cite{sobol}. Inspired by~\cite{QMC_faster_mc}, we select a Sobol sequence which not only shows consistently better performances than the Halton sequence, but also is up to 5 times faster than the MC method, even with lower error rates.

From the view of numerical analysis, an inequality in~\cite{QMC} proves that low-discrepancy sequences guarantees more advanced sampling in~\cref{eq:probability} with fewer integration errors as below:
\begin{equation}
| \alpha | \leqq V(\tau)~D^{*}_N,
\end{equation}
where $V(\tau)$ is a total variation of function $\tau$ which is bounded variation, and $D^{*}_N$ is the discrepancy of a sequence for the number of samples $N$. The inequality shows that a deterministic low-discrepancy sequence can be much better than the random one, for a function with finite variation. 
In the mathematics community, it has been proven that the Sobol sequences have a rate of convergence close to $O((logN)^s/N)$; for a random sequence it is $O(\sqrt{log(logN)/N})$ in~\cite{QMC_faster_mc,QMC}. For faster convergence, $s$ needs to be small and $N$ large (e.g., $N\!>\!2^s$). As a result, the low discrepancy sequences have lower errors for the same number of points ($N\!=\!20$) as shown in \cref{tab:qmc_result}.

\begin{figure}[t]
\centering
\vspace{-1mm}
\includegraphics[width=\columnwidth,trim={0 23mm 0 0},clip]{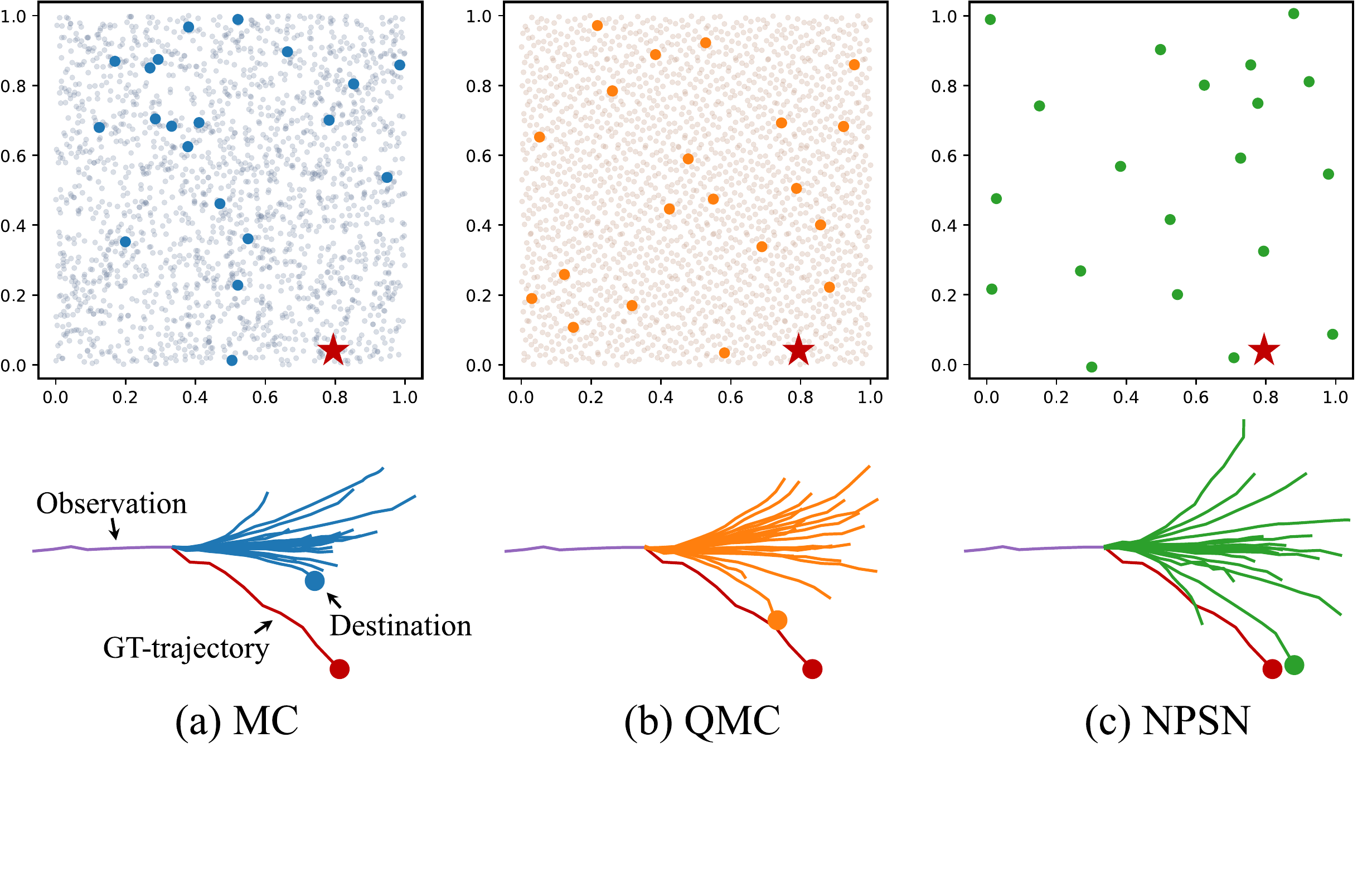}
\vspace{-7mm}
\caption{(Top): 2D scatter plots of 1,000 points with MC, QMC and NPSN. Stars indicate coordinates of a GT destination in the sampling space. (Bottom): Stochastic trajectory prediction results with the first 20 samples of the 1,000 points from each method.}
\vspace{-2mm}
\label{fig:random_points}
\end{figure}

As an example, since $x_i$ are IID samples from a uniformly distributed unit box for MC estimates, the samples tend to be irregularly spaced. For QMC, as $x_i$ comes from a deterministic quasi-random sequence whose point samples are independent, they can be uniformly spaced. This guarantees a suitable distribution for pedestrian trajectory prediction by successively constructing finer uniform partitions. \cref{fig:random_points} displays a plot of a moderate number of pseudo-random points in 2-dimensional space. We observe regions of empty space where there are no points generated from the uniform distribution, which produce results skewed towards the specific destinations. However, the Sobol sequence yields evenly distributed points to enforce prediction results close to socially-acceptable paths.

Unfortunately, low-discrepancy sequences such as the Sobol sequence are deterministically generated and make the trajectory prediction intractable when representing an uncertainty of pedestrians' movements with various social interactions. Adding randomness into the Sobol sequence by scrambling the sequence's base digits~\cite{scramble_sobol} is a solution to this problem. The resultant sequence retains the advantage of QMC method, even with the same expected value. Accordingly, we utilize the scrambled Sobol sequence to generate pedestrian trajectories to account for the feasibility, the diversity, and the randomness of human behaviors.

\vspace{-0.5mm}
\section{Non-Probability Sampling Network}
\vspace{-0.5mm}
In this section, we propose NPSN, which extends the sampling technique for pedestrian trajectory prediction based on observed trajectory. Unlike the previous methods, which sample $N$ paths in a stochastic manner, we construct a model that effectively chooses target samples using a non-probabilistic sampling technique illustrated in~\cref{fig:noise_models}-(d). 

\vspace{-0.5mm}
\subsection{Non-Probability Sampling on Multimodal Trajectory Prediction}
\vspace{-0.5mm}
In contrast to stochastic sampling, purposive sampling, one of the most common non-probability sampling techniques~\cite{black2019business}, relies on the subjective judgment of an expert to select the most productive samples rather than random selection. This approach is advantageous when studying complicated phenomena in in-depth qualitative research~\cite{samplingforqualitative}.

Since most people walk to their destinations using the shortest path, a large portion of labeled attributes in public datasets~\cite{5459260, crowdsbyexample} are straight paths. Generative attribute models learn the probabilistic distributions of social affinity features for the attribute of straight paths. However, due to the multimodal nature of human paths, the models must generate as many diverse and feasible paths as possible, using only a fixed number of samples. As a possible solution, we can purposively include a variety of samples on turning left/right and detouring around obstacles. In purposive sampling, a maximum variation is beneficial for multimodal trajectory prediction, when examining the diverse ranges of pedestrians' movements. We make this process a learnable method, aiming to generate $N$ heterogeneous trajectory samples with prior knowledge of past trajectories.

\subsection{NPSN Architecture}
We propose NPSN which substitutes the random sampling process of existing models with a learnable method. NPSN works as purposive sampling, which relies on the past trajectories of pedestrians when selecting samples in the distribution. 
As a result, when predicting a feasible future trajectory, a past trajectory can be used for the sampling process while also embedding informative features as a guidance. Unlike existing works~\cite{mangalam2020pecnet, li2019idl, salzmann2020trajectron++, yuan2021agent} that impose a restriction in the sampling space by limiting a distribution, we design all of the processes in a learnable manner.

\noindent\textbf{Pedestrian graph representation.}\quad
NPSN first captures the social relations using a GAT to generate socially-acceptable samples. For input trajectory $\smash{X_{l}^{1:T_{obs}}}$, a pedestrian graph $\mathcal{G}\!=\!(\mathcal{V}, \mathcal{E})$ is defined as a set of pedestrian nodes $\mathcal{V}\!=\!\{ v_l\,|\,l\!\in\![1, ..., L] \}$ and their relation edges $\mathcal{E}\!=\! \{ e_{i,j}\,|\,i, j\!\in\![1, ..., L] \}$. With the node features $H\!=\!\{ h_l | l \in [1, ..., L] \}$, learned feature maps for the social relation are shared across different pedestrian nodes in a scene. We utilize an attention mechanism for modeling the social interaction, whose effectiveness is demonstrated in previous works~\cite{huang2019stgat,Shi2021sgcn}. The GAT allows NPSN to aggregate the features for neighbors by assigning different importance to their edge $e_{i,j}$. Here, the importance value is calculated using the attention score between two node features $(h_i, h_j)$.

\noindent\textbf{Purposive sampling.}\quad
With the interaction-aware node features, we predict $N$ samples for each pedestrian. In particular, we use three MLP layers after the GAT layer for NPSN. By learning more prior information about samples of interest, prediction models using NPSN generate better samples. Each trajectory prediction model additionally receives an $s$-dimensional random latent vector along with the observed trajectory. Therefore, the NPSN must predict a set of output $S_l\!=\![S_{l, 1}, ..., S_{l, N}]$. The output passes through a prediction model to generate $N$ final trajectories for each pedestrian. For temporal consistency, we use the same set of purposive samples for all prediction time frames $[1, ..., T_{pred}]$ of each pedestrian node. This process is repeated for all pedestrian nodes, and the output shape of the NPSN is $S=\mathbb{R}^{L \times s \times N}$.

\noindent\textbf{Loss function.}\quad
To optimize trajectory prediction models with our NPSN, we use two loss functions to generate well-distributed purposive samples. First, a winner-takes-all process~\cite{rupprecht2017learning}, which generates a path closest to its ground truth, is trained to regress the accurate positions of pedestrians. Similar to~\cite{gupta2018social}, we measure a $L_2$ distance between the $N$ prediction paths and the ground-truth, and use only one path with the smallest error for training:
\noindent\vspace{-2mm}
\begin{equation}
    \mathcal{L}_{dist} = \frac{1}{L} \sum_{l=1}^{L}\min_{n \in [1, ..., N]} || \hat{Y}_{l, n}^{1:T_{pred}} - Y_{l}^{1:T_{pred}} ||.
    \vspace{-2mm}
\end{equation}

\begin{table*}[t]
    \vspace{-1mm}
    \large
    \centering
    \begin{tabular}{@{}c@{}}
        \resizebox{\linewidth}{!}{
        \begin{tabular}{c ccc cccc cccc c ccc cccc cccc c}
\toprule
\multirow{3}{*}{} & \multicolumn{11}{c}{Social-STGCNN~\cite{mohamed2020social}~~($s=2$)} & & \multicolumn{10}{c}{SGCN~\cite{Shi2021sgcn}~~($s=2$)} \\ \cmidrule(r){2-12} \cmidrule(r){14-24}
& \multicolumn{3}{c}{MC (Baseline)} & \multicolumn{4}{c}{QMC} & \multicolumn{4}{c}{\textbf{NPSN}} & & \multicolumn{3}{c}{MC (Baseline)} & \multicolumn{4}{c}{QMC} & \multicolumn{4}{c}{\textbf{NPSN}} \\ \cmidrule(r){2-4} \cmidrule(lr){5-8} \cmidrule(lr){9-12} \cmidrule(r){14-16} \cmidrule(lr){17-20} \cmidrule(lr){21-24} 
& ADE \darrow & FDE \darrow & TCC \uarrow & ADE \darrow & FDE \darrow & TCC \uarrow & Gain \uarrow & ADE \darrow & FDE \darrow & TCC \uarrow & Gain \uarrow & & ADE \darrow & FDE \darrow & TCC \uarrow & ADE \darrow & FDE \darrow & TCC \uarrow & Gain \uarrow & ADE \darrow & FDE \darrow & TCC \uarrow & Gain \uarrow \\ \midrule
ETH~~~   & 0.650 & 1.097       & 0.510 & \tul{0.611} & \tul{1.025} & \tbf{0.579} & 6.5\%  & \tbf{0.443} & \tbf{0.652} & \tul{0.565} & 40.6\% & & 0.567 & 0.997 & 0.545 & \tul{0.495} & \tul{0.810} & \tul{0.596} & 18.8\% & \tbf{0.357} & \tbf{0.588} & \tbf{0.624} & 41.0\% \\
HOTEL~~~ & 0.496 & 0.858       & 0.270 & \tul{0.342} & \tul{0.517} & \tul{0.289} & 39.8\% & \tbf{0.213} & \tbf{0.342} & \tbf{0.298} & 60.2\% & & 0.308 & 0.533 & 0.295 & \tul{0.212} & \tul{0.309} & \tul{0.314} & 42.0\% & \tbf{0.159} & \tbf{0.253} & \tbf{0.355} & 52.6\% \\
UNIV~~~  & 0.441 & 0.798       & 0.637 & \tul{0.364} & \tul{0.628} & \tul{0.725} & 21.3\% & \tbf{0.278} & \tbf{0.443} & \tbf{0.762} & 44.5\% & & 0.374 & 0.668 & 0.689 & \tul{0.310} & \tul{0.555} & \tul{0.737} & 16.9\% & \tbf{0.229} & \tbf{0.394} & \tbf{0.820} & 41.0\% \\
ZARA1~~~ & 0.341 & 0.532       & 0.710 & \tul{0.315} & \tul{0.526} & \tul{0.775} & 1.1\%  & \tbf{0.248} & \tbf{0.430} & \tbf{0.802} & 19.1\% & & 0.285 & 0.508 & 0.746 & \tul{0.245} & \tul{0.446} & \tul{0.803} & 12.1\% & \tbf{0.182} & \tbf{0.318} & \tbf{0.854} & 37.3\% \\
ZARA2~~~ & 0.305 & \tul{0.482} & 0.394 & \tul{0.288} & 0.497       & \tbf{0.467} & -3.2\% & \tbf{0.217} & \tbf{0.379} & \tul{0.439} & 21.4\% & & 0.225 & 0.422 & 0.491 & \tul{0.193} & \tul{0.359} & \tul{0.503} & 14.9\% & \tbf{0.138} & \tbf{0.245} & \tbf{0.735} & 41.8\% \\ \midrule
AVG~~~   & 0.447 & 0.753       & 0.504 & \tul{0.384} & \tul{0.639} & \tul{0.567} & 15.2\% & \tbf{0.280} & \tbf{0.449} & \tbf{0.573} & 37.2\% & & 0.352 & 0.626 & 0.553 & \tul{0.291} & \tul{0.496} & \tul{0.591} & 20.7\% & \tbf{0.213} & \tbf{0.360} & \tbf{0.678} & 42.5\% \\ 
SDD~~~   & 20.76 & 33.18       & 0.471 & \tul{19.21} & \tul{31.81} & \tul{0.498} & 4.1\%  & \tbf{11.80} & \tbf{18.43} & \tbf{0.551} & 44.5\% & & 25.00 & 41.52 & 0.570 & \tul{21.97} & \tul{38.04} & \tul{0.604} & 8.4\%  & \tbf{17.12} & \tbf{28.97} & \tbf{0.650} & 30.2\% \\ 
GCS~~~   & 14.72 & 23.87       & 0.698 & \tul{13.42} & \tul{22.18} & \tul{0.724} & 7.1\%  & \tbf{9.72}  & \tbf{15.69} & \tbf{0.760} & 34.3\% & & 11.18 & 20.65 & 0.777 & \tul{10.10} & \tul{18.69} & \tbf{0.795} & 9.5\%  & \tbf{7.66}  & \tbf{13.41} & \tul{0.789} & 35.1\% \\ \bottomrule
    \end{tabular}
        } \\ \vspace{-4mm} \\
        \resizebox{\linewidth}{!}{
        \begin{tabular}{c ccc cccc cccc c ccc cccc cccc c}
\toprule
\multirow{3}{*}{} & \multicolumn{11}{c}{Social-GAN~\cite{gupta2018social}~~($s=8$)} & & \multicolumn{10}{c}{STGAT~\cite{huang2019stgat}~~($s=16$)} \\ \cmidrule(r){2-12} \cmidrule(r){14-24}
& \multicolumn{3}{c}{MC (Baseline)} & \multicolumn{4}{c}{QMC} & \multicolumn{4}{c}{\textbf{NPSN}} & & \multicolumn{3}{c}{MC (Baseline)} & \multicolumn{4}{c}{QMC} & \multicolumn{4}{c}{\textbf{NPSN}} \\ \cmidrule(r){2-4} \cmidrule(lr){5-8} \cmidrule(lr){9-12} \cmidrule(r){14-16} \cmidrule(lr){17-20} \cmidrule(lr){21-24} 
& ADE \darrow & FDE \darrow & TCC \uarrow & ADE \darrow & FDE \darrow & TCC \uarrow & Gain \uarrow & ADE \darrow & FDE \darrow & TCC \uarrow & Gain \uarrow & & ADE \darrow & FDE \darrow & TCC \uarrow & ADE \darrow & FDE \darrow & TCC \uarrow & Gain \uarrow & ADE \darrow & FDE \darrow & TCC \uarrow & Gain \uarrow \\ \midrule
ETH~~~   & 0.767 & 1.397 & \tul{0.592} & \tul{0.760} & \tul{1.379} & \tbf{0.596} & 1.3\% & \tbf{0.718} & \tbf{1.264} & 0.539       & 9.5\%  & & 0.679       & 1.203       & \tul{0.576} & \tul{0.668} & \tul{1.175} & \tul{0.576} & 2.4\%  & \tbf{0.612} & \tbf{1.020} & \tbf{0.602} & 15.2\% \\
HOTEL~~~ & 0.434 & 0.876 & \tul{0.322} & \tul{0.431} & \tul{0.870} & \tbf{0.323} & 0.7\% & \tbf{0.385} & \tbf{0.720} & 0.311       & 17.8\% & & 0.346       & \tul{0.661} & \tul{0.338} & \tul{0.342} & 0.663       & \tbf{0.351} & -0.3\% & \tbf{0.308} & \tbf{0.566} & 0.301       & 14.4\% \\
UNIV~~~  & 0.745 & 1.497 & 0.686       & \tul{0.744} & \tul{1.494} & \tul{0.690} & 0.2\% & \tbf{0.711} & \tbf{1.427} & \tbf{0.715} & 4.7\%  & & 0.545       & 1.164       & 0.759       & \tul{0.541} & \tul{1.152} & \tul{0.763} & 1.0\%  & \tbf{0.535} & \tbf{1.133} & \tbf{0.795} & 2.6\%  \\
ZARA1~~~ & 0.346 & 0.693 & \tul{0.801} & \tbf{0.343} & \tul{0.686} & \tbf{0.805} & 0.9\% & \tul{0.344} & \tbf{0.683} & 0.798       & 1.4\%  & & 0.345       & 0.687       & 0.791       & \tul{0.343} & \tul{0.683} & \tul{0.818} & 0.6\%  & \tbf{0.338} & \tbf{0.678} & \tbf{0.822} & 1.4\%  \\
ZARA2~~~ & 0.356 & 0.721 & 0.474       & \tul{0.354} & \tul{0.716} & \tul{0.476} & 0.7\% & \tbf{0.345} & \tbf{0.696} & \tbf{0.491} & 3.5\%  & & \tul{0.304} & \tul{0.620} & \tul{0.508} & \tbf{0.288} & \tbf{0.588} & 0.484       & 5.2\%  & \tul{0.304} & 0.621       & \tbf{0.557} & -0.1\% \\ \midrule
AVG~~~   & 0.530 & 1.037 & \tul{0.575} & \tul{0.526} & \tul{1.029} & \tbf{0.578} & 0.8\% & \tbf{0.501} & \tbf{0.958} & 0.571       & 7.6\%  & & 0.444       & 0.867       & 0.594       & \tul{0.436} & \tul{0.852} & \tul{0.598} & 1.7\%  & \tbf{0.419} & \tbf{0.804} & \tbf{0.616} & 7.3\%  \\ 
SDD~~~   & 13.58 & 24.59 & 0.598       & \tul{13.41} & \tul{24.24} & \tul{0.601} & 1.4\% & \tbf{13.03} & \tbf{23.04} & \tbf{0.630} & 6.3\%  & & 14.85       & 28.17       & 0.590       & \tul{14.82} & \tul{28.12} & \tul{0.594} & 0.2\%  & \tbf{13.67} & \tbf{25.24} & \tbf{0.613} & 10.4\% \\ 
GCS~~~   & 15.85 & 32.57 & 0.783       & \tul{15.80} & \tul{32.44} & \tul{0.785} & 0.4\% & \tbf{15.78} & \tbf{32.17} & \tbf{0.798} & 1.2\%  & & \tul{15.57} & \tul{31.82} & \tbf{0.798} & \tbf{15.55} & \tbf{31.80} & \tbf{0.798} & 0.1\%  & 15.71       & 32.12       & \tbf{0.798} & -0.9\% \\ \bottomrule
    \end{tabular}
         } \\ \vspace{-4mm} \\
        \resizebox{\linewidth}{!}{
        \begin{tabular}{c ccc cccc cccc c ccc cccc cccc c}
\toprule
\multirow{3}{*}{} & \multicolumn{11}{c}{Trajectron++~\cite{salzmann2020trajectron++}~~($s=25$)} & & \multicolumn{10}{c}{PECNet~\cite{mangalam2020pecnet}~~($s=16$)} \\ \cmidrule(r){2-12} \cmidrule(r){14-24}
& \multicolumn{3}{c}{MC (Baseline)} & \multicolumn{4}{c}{QMC} & \multicolumn{4}{c}{\textbf{NPSN}} & & \multicolumn{3}{c}{MC (Baseline)} & \multicolumn{4}{c}{QMC} & \multicolumn{4}{c}{\textbf{NPSN}} \\ \cmidrule(r){2-4} \cmidrule(lr){5-8} \cmidrule(lr){9-12} \cmidrule(r){14-16} \cmidrule(lr){17-20} \cmidrule(lr){21-24} 
& ADE \darrow & FDE \darrow & TCC \uarrow & ADE \darrow & FDE \darrow & TCC \uarrow & Gain \uarrow & ADE \darrow & FDE \darrow & TCC \uarrow & Gain \uarrow & & ADE \darrow & FDE \darrow & TCC \uarrow & ADE \darrow & FDE \darrow & TCC \uarrow & Gain \uarrow & ADE \darrow & FDE \darrow & TCC \uarrow & Gain \uarrow \\ \midrule
ETH~~~   & 0.610 & 1.028 & 0.495       & \tul{0.591} & \tul{0.995} & \tbf{0.503} & 3.2\% & \tbf{0.518} & \tbf{0.780} & \tul{0.499} & 24.1\% & & 0.610       & 1.073 & 0.596       & \tul{0.601} & \tul{1.036} & \tul{0.602} & 3.5\% & \tbf{0.550} & \tbf{0.882} & \tbf{0.618} & 17.8\% \\
HOTEL~~~ & 0.196 & 0.284 & \tul{0.323} & \tul{0.193} & \tul{0.277} & 0.319       & 2.5\% & \tbf{0.157} & \tbf{0.266} & \tbf{0.352} & 6.2\%  & & 0.222       & 0.390 & 0.335       & \tul{0.214} & \tul{0.369} & \tul{0.336} & 5.3\% & \tbf{0.188} & \tbf{0.288} & \tbf{0.359} & 26.1\% \\
UNIV~~~  & 0.304 & 0.545 & \tul{0.765} & \tul{0.297} & \tul{0.531} & \tbf{0.767} & 2.6\% & \tbf{0.266} & \tbf{0.443} & \tul{0.765} & 18.8\% & & 0.335       & 0.558 & 0.752       & \tul{0.326} & \tul{0.533} & \tul{0.759} & 4.5\% & \tbf{0.289} & \tbf{0.439} & \tbf{0.765} & 21.4\% \\
ZARA1~~~ & 0.241 & 0.413 & 0.764       & \tul{0.235} & \tul{0.401} & \tul{0.766} & 2.8\% & \tbf{0.191} & \tbf{0.358} & \tul{0.844} & 13.4\% & & 0.250       & 0.448 & 0.808       & \tul{0.241} & \tul{0.425} & \tul{0.816} & 5.1\% & \tbf{0.209} & \tbf{0.333} & \tbf{0.839} & 25.8\% \\
ZARA2~~~ & 0.175 & 0.319 & 0.639       & \tul{0.170} & \tul{0.307} & \tul{0.641} & 3.5\% & \tbf{0.161} & \tbf{0.278} & \tbf{0.684} & 12.6\% & & 0.186       & 0.332 & 0.596       & \tul{0.178} & \tul{0.310} & \tul{0.609} & 6.7\% & \tbf{0.159} & \tbf{0.252} & \tbf{0.641} & 24.0\% \\ \midrule
AVG~~~   & 0.305 & 0.518 & 0.597       & \tul{0.297} & \tul{0.502} & \tul{0.599} & 3.0\% & \tbf{0.258} & \tbf{0.425} & \tbf{0.629} & 17.9\% & & 0.321       & 0.560 & 0.617       & \tul{0.312} & \tul{0.535} & \tul{0.624} & 4.6\% & \tbf{0.279} & \tbf{0.439} & \tbf{0.644} & 21.7\% \\ 

SDD~~~   & 11.40 & 20.12 & 0.652       & \tul{11.22} & \tul{19.69} & \tbf{0.656} & 2.1\% & \tbf{11.12} & \tbf{18.95} & \tul{0.653} & 5.8\%  & & 9.97        & 15.89 & 0.647       & \tul{9.72}  & \tul{15.22} & \tul{0.652} & 4.2\% & \tbf{8.56}  & \tbf{11.85} & \tbf{0.665} & 25.4\% \\ 
GCS~~~   & 12.75 & 24.23 & \tul{0.802} & \tul{12.47} & \tul{23.50} & \tul{0.802} & 3.0\% & \tbf{12.36} & \tbf{22.98} & \tbf{0.805} & 5.2\%  & & \tul{17.08} & 29.30 & 0.708       & 17.09       & \tul{29.23} & \tul{0.711} & 0.2\% & \tbf{10.13} & \tbf{17.36} & \tbf{0.717} & 40.8\% \\ \bottomrule
    \end{tabular}
        } \\
    \end{tabular}
    \vspace{-2mm}
    \caption{Comparison results of MC, QMC and NPSN  w.r.t. STGCNN~\cite{mohamed2020social}, SGCN~\cite{Shi2021sgcn}, SGAN~\cite{gupta2018social}, STGAT~\cite{huang2019stgat}, Trajectron++~\cite{salzmann2020trajectron++}, and PECNet~\cite{mangalam2020pecnet} in $N\!=\!20$. Models are evaluated on the ETH~\cite{5459260}, UCY~\cite{crowdsbyexample}, SDD~\cite{robicquet2016learning}, and GCS~\cite{yi2015understanding} datasets. (Gain: performance improvement w.r.t. FDE over the baseline models, Unit for ADE and FDE: meter, \textbf{Bold}:Best, \underline{Underline}:Second best)}
    \vspace{-3mm}
    \label{tab:qmc_result}
\end{table*}

However, we observe that all $N$ sample points are sometimes closely located near its ground-truth as learning progresses. 
This is a common problem in purposive sampling, because certain samples can be over-biased due to data imbalance, \ie a large portion of the trajectory moving along one direction of the walkway. For this reason, we introduce a novel discrepancy loss to keep the $N$ sample points with low-discrepancy, as below:
\noindent\vspace{-2mm}
\begin{equation}
    \mathcal{L}_{disc} = \frac{1}{LN} \sum_{l=1}^{L}\sum_{i=1}^{N} -\log \min_{\substack{j \in [1, ..., N] \\ j \neq i}} || S_{l, i} - S_{l, j} ||.
    \vspace{-2mm}
\end{equation}
The objective of discrepancy loss is to maximize distances among the closest neighbors of $N$ samples. If the distance is closer, the loss imposes a higher penalty to ensure their uniform coverage of the sampling space.

The final loss function is a linear combination of both the distance and the discrepancy loss $\mathcal{L} = \mathcal{L}_{dist} + \lambda \mathcal{L}_{disc}$. We set $\lambda=1e\!-\!2$ to balance the scale of both terms.

\subsection{Implementation Details}
\vspace{-0.5mm}
\noindent\textbf{Transformation of one distribution to another.}\quad
While most human trajectory prediction models use a normal distribution, the Sobol sequence and our NPSN are designed to produce a uniform distribution. We bridge the gap by transforming between the uniform distribution and the normal distribution. There are some representative methods including Ziggurat method~\cite{JSSv005i08}, Inverse Cumulative Distribution Function (ICDF), and Box-Muller Transform~\cite{Box1958ANO}. 
In this work, we utilize the Box-Muller transform which is differentiable and enables an efficient execution on a GPU with the lowest QMC error, as demonstrated in~\cite{gpugems3,OKTEN20111268}.
The formula of the Box-muller transform is as follows:
\noindent
\begin{equation}
\begin{split}
Z_{odd}  &= \sqrt{-2\;\!\:\!ln\:\!(U_{even})}~cos(2\:\!\pi\:\!U_{odd}), \\
Z_{even} &= \sqrt{-2\;\!\:\!ln\:\!(U_{even})}~sin(2\:\!\pi\:\!U_{odd}), 
\end{split}
\end{equation}
where $U$ is an independent sample set from a uniform distribution and $Z$ is an independent random variable from a standard normal distribution.

\noindent\textbf{Training Procedure.}\quad
Our NPSN is embedded into the state-of-the-art pedestrian trajectory prediction models~\cite{mohamed2020social, Shi2021sgcn, gupta2018social, huang2019stgat, salzmann2020trajectron++, mangalam2020pecnet,liu2020snce,liu2021causal} by simply replacing their random sampling part. The parameters of the models are initialized using the weights provided by the authors, except for four models~\cite{huang2019stgat, mangalam2020pecnet, salzmann2020trajectron++, liu2020snce} which use weights reproduced from the authors' source codes. Our NPSN has only 5,128 learnable parameters on $s\!=\!2$ and $N\!=\!20$. We train the prediction models with NPSN using an AdamW optimizer~\cite{loshchilov2018decoupled} with a batch size of 128 and a learning rate of $1e\!-\!3$ for 128 epochs. We step down the learning rate with a gain of 0.5 at every 32 epochs. Training time takes about three hours on a machine with an NVIDIA 2080TI GPU.

\vspace{-1mm}
\section{Experiments}
\vspace{-1mm}
In this section, we conduct comprehensive experiments on public benchmark datasets to verify how the sampling strategy contributes to pedestrian trajectory prediction.
We first briefly describe our experimental setup (\cref{subsec:ExperimentalSetup}), and then provide comparison results with various baselines and state-of-the-art models (\cref{subsec:method}). Moreover, we run an extensive ablation study to demonstrate the effect of each component of our method (\cref{subsec:ablation}).

\vspace{-1mm}
\subsection{Experimental Setup}
\vspace{-1mm}
\label{subsec:ExperimentalSetup}
\noindent\textbf{Dataset.}\quad
We evaluate the effectiveness of the QMC method and our NPSN on various benchmark datasets~\cite{5459260, crowdsbyexample, robicquet2016learning,yi2015understanding } over state-of-the-art methods. ETH~\cite{5459260} and UCY dataset~\cite{crowdsbyexample} include ETH and HOTEL, and UNIV, ZARA1 and ZARA2 scenes, respectively. Both datasets consist of various movements of pedestrians with complicated social interactions. The Stanford Drone Dataset (SDD)~\cite{robicquet2016learning} contains secluded scenes with various object types (\eg pedestrian, biker, skater, and cart), and the Grand Central Station (GCS)~\cite{yi2015understanding} dataset consists of highly congested scenes where pedestrians walk. We observe a trajectory for 3.2 seconds ($T_{obs}\!=\!8$), and then predict future paths for the next 4.8 seconds ($T_{pred}\!=\!12$). We follow a leave-one-out cross-validation evaluation strategy, which is the standard evaluation protocol used in many works~\cite{gupta2018social, huang2019stgat, mohamed2020social, Shi2021sgcn, salzmann2020trajectron++, mangalam2020pecnet}.

\noindent\textbf{Evaluation metric.}\quad
We measure the performance of the trajectory prediction models using three metrics: 1) Average Displacement\,Error\,(ADE) - average\,Euclidean\,distance between a prediction and ground-truth trajectory; 2) Final Displacement Error (FDE) - Euclidean distance between a prediction and ground-truth final destination; 3) Temporal Correlation Coefficient (TCC)~\cite{tao2020dynamic} - Pearson correlation coefficient of motion patterns between a prediction and ground-truth trajectory. These metrics assess the best one of $N\!=\!20$ trajectory outputs, and we report average values for all agents in each scene. In addition, to reduce the variance in the prediction results of stochastic models, we repeat the evaluation 100 times and then average them for each metric.

\noindent\textbf{Baseline.}\quad
We evaluate QMC and NPSN sampling methods with representative stochastic pedestrian trajectory prediction models:
1) Gaussian distribution-based model - Social-STGCNN\,\cite{mohamed2020social}, SGCN\,\cite{Shi2021sgcn};
2) GAN-based model - Social-GAN\,\cite{gupta2018social}, STGAT\,\cite{huang2019stgat}, Causal-STGAT\,\cite{liu2021causal};
3) CVAE-based model - Trajectron++\,\cite{salzmann2020trajectron++}, PECNet\,\cite{mangalam2020pecnet}, and NCE-Trajectron++\,\cite{liu2020snce}. 
To validate the effectiveness of QMC and NPSN, we replace their random sampling parts in the authors' provided codes with our QMC and NPSN sampling method.

\vspace{-1mm}
\subsection{Results from QMC and NPSN method}
\begin{figure}[t]
\centering
\vspace{-0.5mm}
\includegraphics[width=\linewidth,trim={0 11mm 0 0},clip]{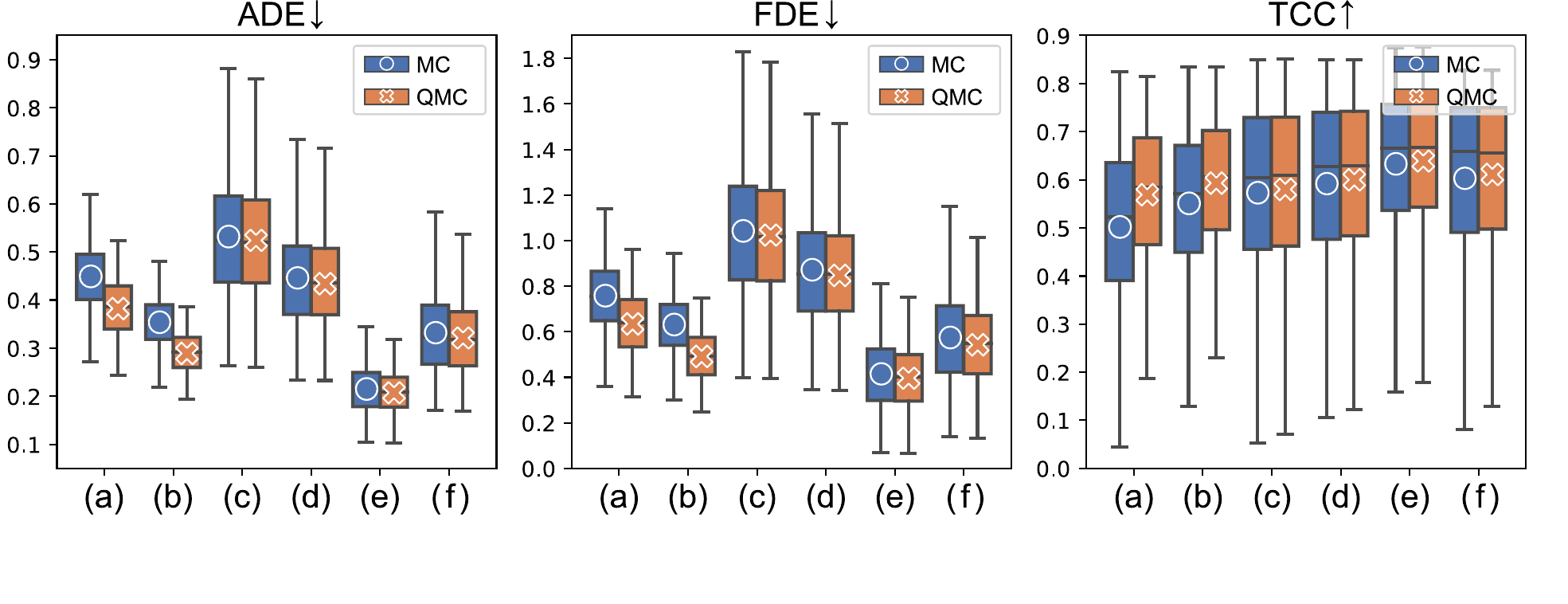}
\vspace{-7mm}
\caption{Box plots of average ADE, FDE and TCC measured for each stochastic model on both MC and QMC. (a) STGCNN~\cite{mohamed2020social}, (b) SGCN~\cite{Shi2021sgcn}, (c) SGAN~\cite{gupta2018social}, (d) STGAT~\cite{huang2019stgat}, (e) Trajectron++~\cite{salzmann2020trajectron++}, and (f) PECNet~\cite{mangalam2020pecnet}.}
\vspace{-2mm}
\label{fig:boxplot}
\end{figure}

\vspace{-1mm}
\label{subsec:method}
\noindent\textbf{Comparison of MC and QMC.}\quad
We compare MC with the QMC method by incorporating them into the sampling part of the baseline models. As shown in~\cref{tab:qmc_result,fig:boxplot}, the QMC method significantly outperforms the MC method on all the evaluation metrics. In~\cref{fig:boxplot}, we report the error distributions of the baseline models in the test phase. The QMC method achieves consistently lower errors and variations by alleviating the bias problem mentioned in~\cref{subsec:bias}.

We also observe that the Gaussian-based models show a large performance gain over the GAN- and CVAE-based models. There are two reasons for the performance gains induced by the QMC method: 
1) The dimension of the sampling space ($s\!=\!2$) in the Gaussian-based models is relatively smaller than other models (\ie $s\!=\!8$, $16$ or $25$). According to~\cite{drawbackQMC}, for large dimensions $s$ and a small number of samples $N$, the sampling results from a low-discrepancy generator may not be good enough over randomly generated samples. The Gaussian-based model thus yields promising results compared to one which has larger sampling dimensions. 
2) The performance improvements depend on the number of layers in networks (shallower is better): The CVAE and GAN-based models are composed of multiple layers. By contrast, the Gaussian-based models have only one layer which acts as a linear transformation between the predicted trajectory coordinates and final coordinates. To be specific, in the transformation, sampled independent 2D points are multiplied with the Cholesky decomposed covariance matrix and shifted by the mean matrix. Here, the shallow layer of the Gaussian-based models directly reflects the goodness of the QMC sampling method, rather than deeper layers which can barely be influenced by the random latent vector in the inference step.

\begin{figure}[t]
\centering
\includegraphics[width=\columnwidth,trim={0 43mm 0 0},clip]{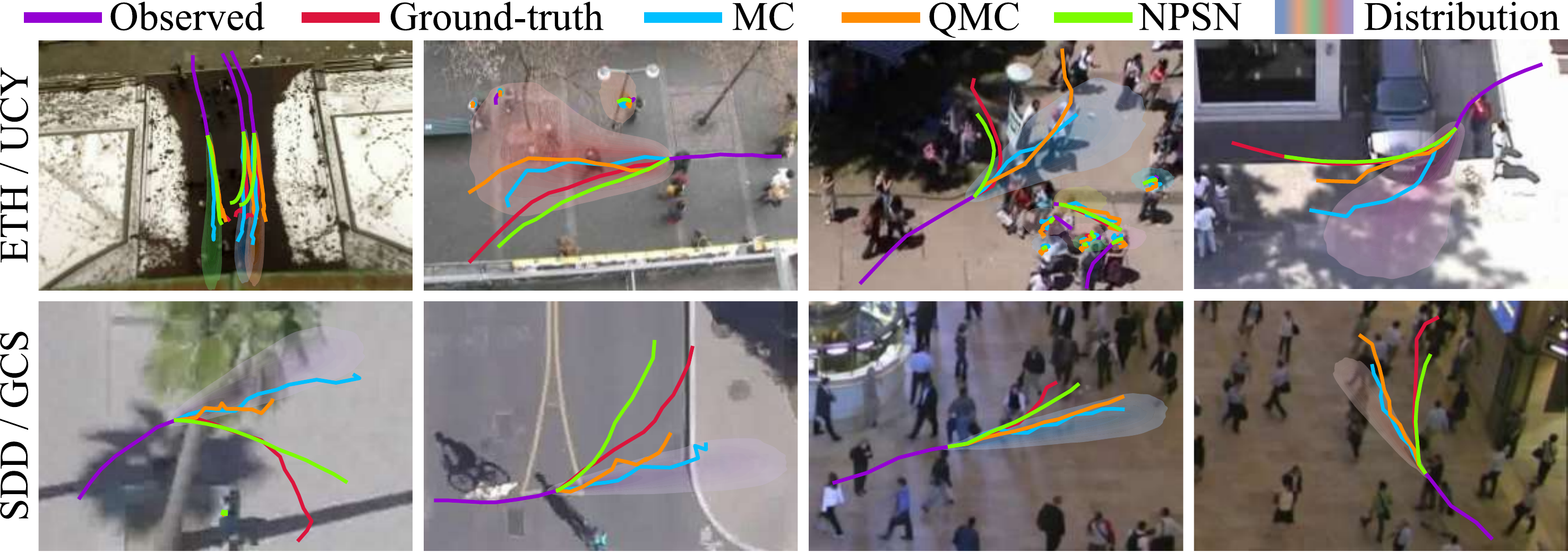}
\vspace{-7mm}
\caption{Visualization of probabilistic distributions and the best predictions among sampled trajectories with MC, QMC, and NPSN in SGCN~\cite{Shi2021sgcn}.}
\vspace{-3mm}
\label{fig:qualitative_result}
\end{figure}

\noindent\textbf{Evaluation of NPSN.}\quad
We apply NPSN to all three types of stochastic trajectory prediction models.
As shown in~\cref{tab:qmc_result}, there are different performance gains according to the types. 
Particularly, the Gaussian distribution approaches (Social-STGCNN~\cite{mohamed2020social}, SGCN~\cite{Shi2021sgcn}) show the highest performance improvement (up to 60\%), which can be analyzed by the advantages of the QMC method when $s=2$. So far, the performance of the Gaussian distribution approaches has been underestimated due to the disadvantage of being easily affected by the sampling bias. Our NPSN maximizes the capability of the Gaussian distribution approaches through a purposive sampling technique.

In the CVAE based approaches, PECNet~\cite{mangalam2020pecnet} shows a larger performance improvement (up to 41\%) than that of Trajectron++~\cite{salzmann2020trajectron++}. Since PECNet directly predicts a set of destinations through the latent vector, NPSN is compatible with its inference step. On the other hand, NPSN seems to produce less benefit with the inference step of Trajectron++ because it predicts the next step recurrently and its sample dimension is relatively large ($s=25$).

The generative models with variety loss, Social-GAN and STGAT, show relatively small performance improvements, compared to the others. For some datasets, the FDE values of STGAT 
are lower than those of MC and QMC when using our NPSN. This seems to suggest that NPSN fails to learn samples close to ground-truth trajectories due to the common entanglement problem of latent space~\cite{infogan,stylegan}.

\begin{table}[t]
    \Large
    \centering
    \resizebox{\linewidth}{!}{
\begin{tabular}{ccccccc}
\toprule
                                             & ETH                     & HOTEL                   & UNIV                    & ZARA1                   & ZARA2                   & AVG                     \\ \midrule
Social-GAN~\cite{gupta2018social}            & 0.87 / 1.62             & 0.67 / 1.37             & 0.76 / 1.52             & 0.35 / 0.68             & 0.42 / 0.84             & 0.61 / 1.21             \\
STGAT~\cite{huang2019stgat}                  & 0.65 / 1.12             & 0.35 / 0.66             & 0.52 / 1.10             & 0.34 / 0.69             & 0.29 / 0.60             & 0.43 / 0.83             \\
Causal-STGAT~\cite{liu2021causal}            & 0.60 / 0.98             & 0.30 / 0.54             & 0.52 / 1.10             & 0.32 / 0.64             & 0.28 / 0.58             & 0.40 / 0.77             \\
Social-STGCNN~\cite{mohamed2020social}       & 0.64 / 1.11             & 0.49 / 0.85             & 0.44 / 0.79             & 0.34 / 0.53             & 0.30 / 0.48             & 0.44 / 0.75             \\
PECNet~\cite{mangalam2020pecnet}             & 0.61 / 1.07             & 0.22 / 0.39             & 0.34 / 0.56             & 0.25 / 0.45             & 0.19 / 0.33             & 0.32 / 0.56             \\
Trajectron++~\cite{salzmann2020trajectron++} & 0.61 / 1.03             & 0.20 / 0.28             & 0.30 / 0.55             & 0.24 / 0.41             & 0.18 / 0.32             & 0.31 / 0.52             \\
NCE-Trajectron++~\cite{liu2020snce}          & 0.56 / 1.02             & 0.17 / 0.27             & 0.28 / 0.54             & 0.22 / 0.41             & \tul{0.16} / 0.31       & 0.28 / 0.51             \\
SGCN~\cite{Shi2021sgcn}                      & 0.57 / 1.00             & 0.31 / 0.53             & 0.37 / 0.67             & 0.29 / 0.51             & 0.22 / 0.42             & 0.35 / 0.63             \\ \cmidrule(lr){1-7}
\tbf{NPSN-SGAN}                              & 0.72 / 1.26             & 0.38 / 0.72             & 0.71 / 1.43             & 0.34 / 0.68             & 0.34 / 0.70             & 0.50 / 0.96             \\
\tbf{NPSN-STGAT}                             & 0.61 / 1.02             & 0.31 / 0.57             & 0.53 / 1.13             & 0.34 / 0.68             & 0.30 / 0.62             & 0.42 / 0.80             \\
\tbf{NPSN-Causal-STGAT}                      & 0.56 / 0.90             & 0.25 / 0.40             & 0.51 / 1.09             & 0.32 / 0.65             & 0.27 / 0.56             & 0.38 / 0.72             \\
\tbf{NPSN-STGCNN}                            & 0.44 / 0.65             & 0.21 / 0.34             & 0.28 / 0.44             & 0.25 / 0.43             & 0.22 / 0.38             & 0.28 / 0.45             \\
\tbf{NPSN-PECNet}                            & 0.55 / 0.88             & 0.19 / 0.29             & 0.29 / 0.44             & 0.21 / \tul{0.33}       & \tul{0.16} / \tbf{0.25} & 0.28 / 0.44             \\
\tbf{NPSN-Trajectron++}                      & 0.52 / 0.78             & \tul{0.16} / 0.27       & \tul{0.27} / 0.44       & \tul{0.19} / 0.36       & \tul{0.16} / \tul{0.28} & 0.26 / 0.42             \\
\tbf{NPSN-NCE-Trajectron++}                  & \tul{0.40} / \tul{0.62} & \tbf{0.15} / \tbf{0.24} & \tbf{0.23} / \tul{0.41} & \tul{0.19} / 0.35       & \tbf{0.14} / \tbf{0.25} & \tul{0.22} / \tul{0.37} \\
\tbf{NPSN-SGCN}                              & \tbf{0.36} / \tbf{0.59} & \tul{0.16} / \tul{0.25} & \tbf{0.23} / \tbf{0.39} & \tbf{0.18} / \tbf{0.32} & \tbf{0.14} / \tbf{0.25} & \tbf{0.21} / \tbf{0.36} \\
\bottomrule
\end{tabular}
}
\vspace{-3mm}
\caption{Comparison of NPSN with other state-of-the-art stochastic model (ADE/FDE, Unit: meter). The evaluation results are directly referred from~\cite{liu2021causal}. \textbf{Bold}:Best, \underline{Underline}:Second Best.}
\vspace{-2mm}
\label{tab:npsn_sota}
\end{table}

\noindent\textbf{Qualitative results.}\quad
\cref{fig:qualitative_result} shows several cases where there are differences between the predictions of NPSN and other methods. Since NPSN takes an observation trajectory along with the low-discrepancy characteristics of the QMC method, the predicted paths from NPSN are closer to socially-acceptable paths compared to other methods. 

As we described in the~\cref{fig:random_points}, the QMC method generates a more realistic trajectory distribution than the MC method. However, due to the limitations of the dataset, the generated trajectories of the baseline network are biased toward a straight path. 
On the other hand, NPSN sampling method alleviates the problem by selecting the point near the ground-truth in the latent space. As a result, the human trajectory model with NPSN not only generates well-distributed samples with finite sampling pathways, but also represents the feasible range of human's movements.

\noindent\textbf{Comparison with the state-of-the-art models.}\quad
We push the state-of-the-art models with our NPSN, a purposive sampling technique. As shown in~\cref{tab:qmc_result}, our NPSN shows a significant performance improvement on all the baseline networks. NPSN provides better overall accuracy by taking fully advantage of the low-discrepancy characteristics of the QMC method. 

In addition, we report a benchmark result on ETH/UCY dataset in~\cref{tab:npsn_sota}. It is noticeable that all the baseline models exhibit better performances with our NPSN. In particular, when NPSN is incorporated into the combinational approach of Trajectron++~\cite{salzmann2020trajectron++} and NCE~\cite{liu2020snce}, it achieves the best performances on the benchmark. Our NPSN is trained to only control the latent vector samples for the baseline models, and synergizes well with the inference step that comes after both the initial prediction of Trajectron++ and the collision avoidance of NCE.

\begin{figure}[t]
\centering
\vspace{-1mm}
\includegraphics[width=\columnwidth,trim={0 17mm 0 0},clip]{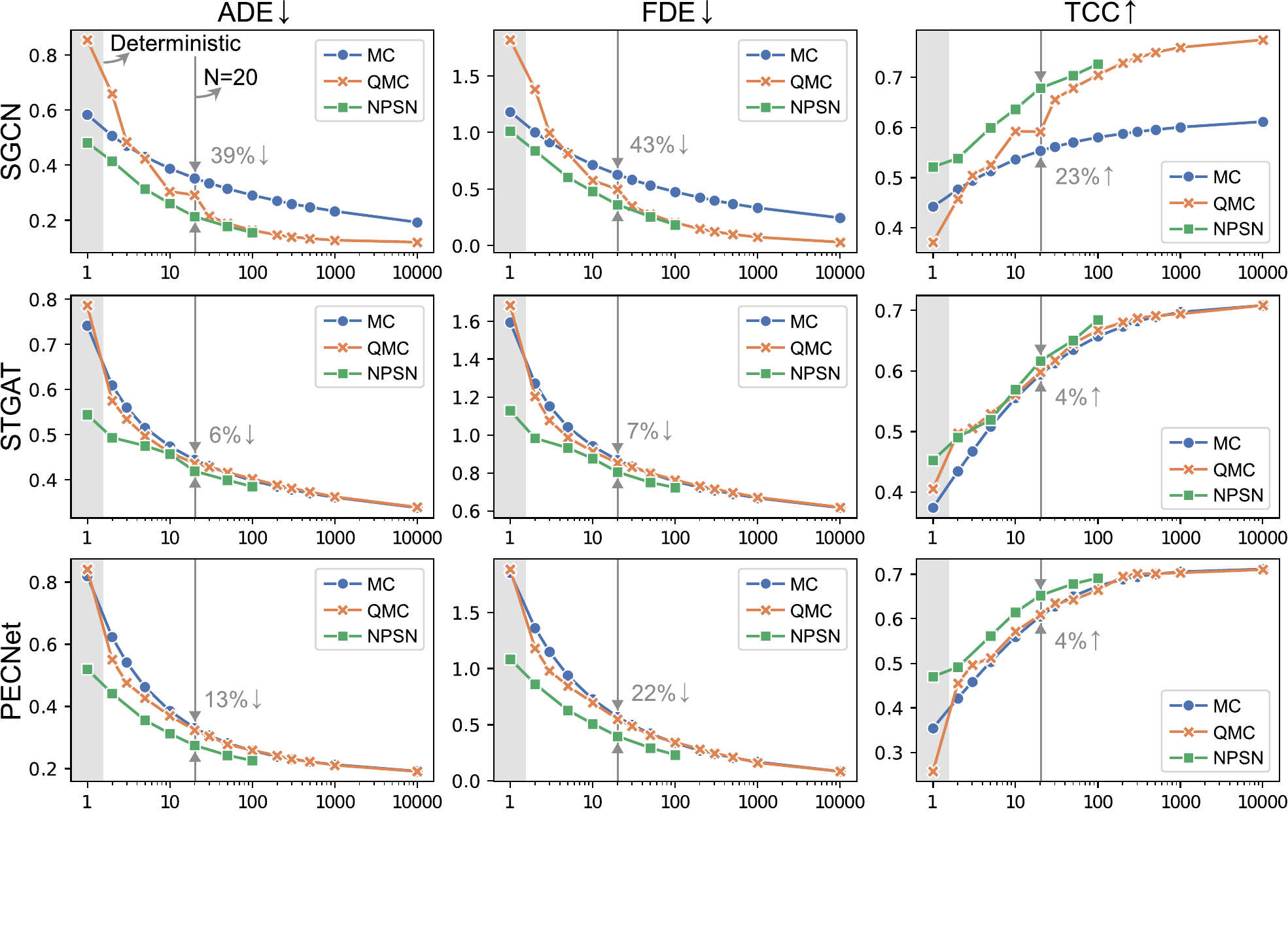}
\vspace{-7mm}
\caption{Averaged ADE/FDE/TCC results on ETH/UCY datasets w.r.t. the number of samples for SGCN~\cite{Shi2021sgcn} (Gaussian distribution approach), STGAT~\cite{huang2019stgat} (GAN-based approach) and PECNet~\cite{mangalam2020pecnet} (CVAE-based approach). We quantify the performance change w.r.t. the sampling methods including MC, QMC and NSPN. We also report the results of deterministic trajectory prediction when $N\!=\!1$ in gray colored regions.}
\vspace{-3.5mm}
\label{fig:graph_change_N}
\end{figure}

\vspace{-0.5mm}
\subsection{Ablation Studies}
\vspace{-0.5mm}
\label{subsec:ablation}
\noindent\textbf{Evaluation of different number of samples.}\quad
To check the effectiveness of the density of sampled paths in human trajectory prediction, we randomly generate trajectories by changing the number of samples $N$. As shown in~\cref{fig:graph_change_N}, the performance gap between the MC and the QMC method is marginal when the number of samples goes to infinity. As mentioned above, it follows the Strong Law of large numbers in the MC integration. 
The Gaussian-based model, SGCN~\cite{Shi2021sgcn}, achieves superior performance and improves more than 30\% performance gain over the classic policy ($N\!=\!20$). Since the sample dimension is small, the effectiveness and convergence of our NPSN are enlarged. Note that a performance drop over sparse conditions due to the discrepancy property: For small $N$ and a comparably large sample space dimension (\ie, $N < 2^s$), the discrepancy of the QMC method may not be less than that of a random sequence. We overcome these limitations with a learnable sampling method by sampling a feasible latent vector with low-discrepancy characteristics.

\noindent\textbf{Deterministic trajectory prediction.}\quad
Since the stochastic model is trained to predict multi-modal future paths, it outputs diverse paths at each execution, which is undesirable for deterministic human trajectory prediction, which infers only one feasible pathway ($N\!=\!1$). By replacing the conventional probability process with a learnable sampling, NPSN allows the stochastic models to infer the most feasible trajectory in a deterministic manner. As shown in~\cref{fig:graph_change_N} (gray colored regions), NPSN outperforms QMC and the conventional methods on all the metrics at $N\!=\!1$.

\begin{table}[t]
    \vspace{-0.5mm}
    \centering
    \resizebox{\linewidth}{!}{
\begin{tabular}{ccccccc}
\toprule
                         & ETH                     & HOTEL                   & UNIV                    & ZARA1                   & ZARA2                   & AVG                     \\ \midrule
~~~Baseline~~~           & 0.57 / 1.00             & 0.31 / 0.53             & 0.37 / 0.67             & 0.29 / 0.51             & 0.22 / 0.42             & 0.35 / 0.63             \\ \cmidrule(lr){1-7}
w/o $\mathcal{L}_{dist}$ & 0.39 / 0.61             & 0.23 / 0.45             & \tul{0.26} / \tul{0.47} & \tul{0.20} / \tul{0.36} & \tul{0.16} / 0.31       & 0.25 / 0.44             \\
w/o $\mathcal{L}_{disc}$ & \tul{0.38} / 0.61       & \tbf{0.16} / \tbf{0.25} & \tbf{0.23} / \tbf{0.39} & \tbf{0.18} / \tbf{0.32} & \tbf{0.14} / \tbf{0.25} & \tul{0.22} / \tul{0.37} \\
w/o GAT                  & \tbf{0.36} / \tbf{0.57} & \tul{0.17} / \tul{0.28} & \tbf{0.23} / \tbf{0.39} & \tbf{0.18} / \tbf{0.32} & \tbf{0.14} / \tul{0.26} & \tul{0.22} / \tul{0.37} \\
+NPSN                    & \tbf{0.36} / \tul{0.59} & \tbf{0.16} / \tbf{0.25} & \tbf{0.23} / \tbf{0.39} & \tbf{0.18} / \tbf{0.32} & \tbf{0.14} / \tbf{0.25} & \tbf{0.21} / \tbf{0.36} \\ 
\bottomrule
\end{tabular}
}
\vspace{-3mm}
\caption{Ablation study\,on\,each\,component\,of\,NPSN in SGCN.}
\vspace{-5mm}
\label{tab:npsn_ablation}
\end{table}

\noindent\textbf{Effectiveness of each component.}\quad
Lastly, we examine the effectiveness of each component in our NPSN, whose result is reported in~\cref{tab:npsn_ablation}. Here, SGCN~\cite{Shi2021sgcn} is selected as the baseline model because it shows the most significant performance improvements with NPSN. First, our two loss functions work well. Particularly, the discrepancy loss guarantees sample diversity by generating low-discrepancy samples, and the distance loss enforces generating samples close to the ground-truth trajectory. The GAT captures the agent-aware interaction for socially-acceptable trajectory prediction, except for the secluded ETH scene.

\vspace{-0.5mm}
\section{Conclusion}
\vspace{-1.5mm}
In this work, we numerically analyze the limitations of the conventional sampling process in stochastic pedestrian trajectory prediction, by using the concept of discrepancy as a measure of the sampling quality. To overcome this limitation, we then introduce a novel, light-weight and learnable sampling strategy, inspired by the Quasi-Monte Carlo method. Unlike conventional random sampling, our learnable method considers both observations and the social norms of pedestrians in scenes. In addition, our method can be inserted into stochastic pedestrian trajectory predictions as a plug-and-play module. With the proposed learnable method, all of the state-of-art models achieve performance improvements. In particular, the Gaussian-based models show the best results on the benchmark. 

\vspace{3mm}
\fontsize{7.24}{8}\selectfont{\noindent\textbf{Acknowledgement} This work is in part supported by the Institute of Information $\&$ communications Technology Planning $\&$ Evaluation (IITP) grant funded by the Korea government (MSIT) (No.2019-0-01842, Artificial Intelligence Graduate School Program (GIST), No.2021-0-02068, Artificial Intelligence Innovation Hub), Vehicles AI Convergence Research $\&$ Development Program through the National IT Industry Promotion Agency of Korea (NIPA) funded by the Ministry of Science and ICT(No.S1602-20-1001), the National Research Foundation of Korea (NRF) grant funded by the Korea government (MSIT) (No.2020R1C1C1012635), and the GIST-MIT Collaboration grant funded by the GIST in 2022.}

\clearpage
{\small
\bibliographystyle{ieee_fullname}
\bibliography{egbib}

\begin{thebibliography}{10}\itemsep=-1pt

\bibitem{alahi2016social}
Alexandre Alahi, Kratarth Goel, Vignesh Ramanathan, Alexandre Robicquet, Li
  Fei-Fei, and Silvio Savarese.
\newblock Social lstm: Human trajectory prediction in crowded spaces.
\newblock In {\em Proceedings of IEEE Conference on Computer Vision and Pattern
  Recognition (CVPR)}, 2016.

\bibitem{Bae_Jeon_2021}
Inhwan Bae and Hae-Gon Jeon.
\newblock Disentangled multi-relational graph convolutional network for
  pedestrian trajectory prediction.
\newblock {\em Thirty-Fifth AAAI Conference on Artificial Intelligence}, 2021.

\bibitem{scramble_sobol}
Peter Beerli, Deidre Evans, and Michael Mascagni.
\newblock On the scrambled sobol sequence.
\newblock {\em Lecture Notes in Computer Science}, 2005.

\bibitem{black2019business}
Ken Black.
\newblock {\em Business statistics: for contemporary decision making}.
\newblock John Wiley \& Sons, USA, 2019.

\bibitem{Box1958ANO}
George E.~P. Box and Mervin~E. Muller.
\newblock A note on the generation of random normal deviates.
\newblock {\em Annals of Mathematical Statistics}, 1958.

\bibitem{liu2021causal}
Guangyi Chen, Junlong Li, Jiwen Lu, and Jie Zhou.
\newblock Human trajectory prediction via counterfactual analysis.
\newblock In {\em Proceedings of International Conference on Computer Vision
  (ICCV)}, 2021.

\bibitem{infogan}
Xi Chen, Yan Duan, Rein Houthooft, John Schulman, Ilya Sutskever, and Pieter
  Abbeel.
\newblock Infogan: Interpretable representation learning by information
  maximizing generative adversarial nets.
\newblock In {\em Proceedings of the Neural Information Processing Systems
  (NeurIPS)}, 2016.

\bibitem{MC_NAS_1}
Xiaoliang Dai, Alvin Wan, Peizhao Zhang, Bichen Wu, Zijian He, Zhen Wei, Kan
  Chen, Yuandong Tian, Matthew Yu, Peter Vajda, and Joseph~E. Gonzalez.
\newblock Fbnetv3: Joint architecture-recipe search using predictor
  pretraining.
\newblock In {\em Proceedings of IEEE Conference on Computer Vision and Pattern
  Recognition (CVPR)}, 2021.

\bibitem{MC_NAS_2}
Xiaoliang Dai, Peizhao Zhang, Bichen Wu, Hongxu Yin, Fei Sun, Yanghan Wang,
  Marat Dukhan, Yunqing Hu, Yiming Wu, Yangqing Jia, Peter Vajda, Matt
  Uyttendaele, and Niraj~K. Jha.
\newblock Chamnet: Towards efficient network design through platform-aware
  model adaptation.
\newblock In {\em Proceedings of IEEE Conference on Computer Vision and Pattern
  Recognition (CVPR)}, 2019.

\bibitem{dendorfer2021mggan}
Patrick Dendorfer, Sven Elflein, and Laura Leal-Taixé.
\newblock Mg-gan: A multi-generator model preventing out-of-distribution
  samples in pedestrian trajectory prediction.
\newblock In {\em Proceedings of International Conference on Computer Vision
  (ICCV)}, 2021.

\bibitem{MC_3DPointCloudRegist}
Zhi Deng, Yuxin Yao, Bailin Deng, and Juyong Zhang.
\newblock A robust loss for point cloud registration.
\newblock In {\em Proceedings of International Conference on Computer Vision
  (ICCV)}, 2021.

\bibitem{LawofLargeNumber}
William Feller.
\newblock {\em An introduction to probability theory and its applications.
  {V}ol. {II}.}
\newblock John Wiley \& Sons Inc., New York, 1971.

\bibitem{gupta2018social}
Agrim Gupta, Justin Johnson, Li Fei-Fei, Silvio Savarese, and Alexandre Alahi.
\newblock Social gan: Socially acceptable trajectories with generative
  adversarial networks.
\newblock In {\em Proceedings of IEEE Conference on Computer Vision and Pattern
  Recognition (CVPR)}, 2018.

\bibitem{halton}
J.~H. Halton.
\newblock Algorithm 247: Radical-inverse quasi-random point sequence.
\newblock {\em Commun. ACM}, 1964.

\bibitem{helbing1995social}
Dirk Helbing and Peter Molnar.
\newblock Social force model for pedestrian dynamics.
\newblock {\em Physical review E}, 51(5):4282, 1995.

\bibitem{gpugems3}
Lee Howes and David Thomas.
\newblock {\em GPU Gems 3 - Efficient Random Number Generation and Application
  Using CUDA}.
\newblock Pearson Education, Inc, 2007.

\bibitem{huang2019stgat}
Yingfan Huang, Huikun Bi, Zhaoxin Li, Tianlu Mao, and Zhaoqi Wang.
\newblock Stgat: Modeling spatial-temporal interactions for human trajectory
  prediction.
\newblock In {\em Proceedings of International Conference on Computer Vision
  (ICCV)}, 2019.

\bibitem{Ivanovic_2019_ICCV}
Boris Ivanovic and Marco Pavone.
\newblock The trajectron: Probabilistic multi-agent trajectory modeling with
  dynamic spatiotemporal graphs.
\newblock In {\em Proceedings of International Conference on Computer Vision
  (ICCV)}, 2019.

\bibitem{sobol}
Stephen Joe and Frances~Y. Kuo.
\newblock Constructing sobol sequences with better two-dimensional projections.
\newblock {\em SIAM Journal on Scientific Computing}, 2008.

\bibitem{stylegan}
Tero Karras, Samuli Laine, and Timo Aila.
\newblock A style-based generator architecture for generative adversarial
  networks.
\newblock In {\em Proceedings of IEEE Conference on Computer Vision and Pattern
  Recognition (CVPR)}, 2019.

\bibitem{kipf2016semi}
Thomas~N Kipf and Max Welling.
\newblock Semi-supervised classification with graph convolutional networks.
\newblock In {\em International Conference on Learning Representations (ICLR)},
  2017.

\bibitem{kosaraju2019social}
Vineet Kosaraju, Amir Sadeghian, Roberto Mart{\'\i}n-Mart{\'\i}n, Ian Reid,
  Hamid Rezatofighi, and Silvio Savarese.
\newblock Social-bigat: Multimodal trajectory forecasting using bicycle-gan and
  graph attention networks.
\newblock In {\em Proceedings of the Neural Information Processing Systems
  (NeurIPS)}, 2019.

\bibitem{Lee_2017_CVPR}
Namhoon Lee, Wongun Choi, Paul Vernaza, Christopher~B. Choy, Philip H.~S. Torr,
  and Manmohan Chandraker.
\newblock Desire: Distant future prediction in dynamic scenes with interacting
  agents.
\newblock In {\em Proceedings of IEEE Conference on Computer Vision and Pattern
  Recognition (CVPR)}, 2017.

\bibitem{drawbackQMC}
Christiane Lemieux.
\newblock {\em Monte Carlo and Quasi-Monte Carlo Sampling}.
\newblock Springer, Dordrecht, 2009.

\bibitem{crowdsbyexample}
Alon Lerner, Yiorgos Chrysanthou, and Dani Lischinski.
\newblock Crowds by example.
\newblock {\em Computer Graphics Forum}, 26(3):655--664, 2007.

\bibitem{li2019conditional}
Jiachen Li, Hengbo Ma, and Masayoshi Tomizuka.
\newblock Conditional generative neural system for probabilistic trajectory
  prediction.
\newblock {\em Proceedings of IEEE International Conference on Intelligent
  Robots and Systems (IROS)}, 2019.

\bibitem{li2020Evolvegraph}
Jiachen Li, Fan Yang, Masayoshi Tomizuka, and Chiho Choi.
\newblock Evolvegraph: Multi-agent trajectory prediction with dynamic
  relational reasoning.
\newblock In {\em Proceedings of the Neural Information Processing Systems
  (NeurIPS)}, 2020.

\bibitem{li2019idl}
Yuke Li.
\newblock Which way are you going? imitative decision learning for path
  forecasting in dynamic scenes.
\newblock In {\em Proceedings of IEEE Conference on Computer Vision and Pattern
  Recognition (CVPR)}, 2019.

\bibitem{liang2020simaug}
Junwei Liang, Lu Jiang, and Alexander Hauptmann.
\newblock Simaug: Learning robust representations from simulation for
  trajectory prediction.
\newblock In {\em Proceedings of European Conference on Computer Vision
  (ECCV)}, 2020.

\bibitem{liang2020garden}
Junwei Liang, Lu Jiang, Kevin Murphy, Ting Yu, and Alexander Hauptmann.
\newblock The garden of forking paths: Towards multi-future trajectory
  prediction.
\newblock In {\em Proceedings of IEEE Conference on Computer Vision and Pattern
  Recognition (CVPR)}, 2020.

\bibitem{liang2019peeking}
Junwei Liang, Lu Jiang, Juan~Carlos Niebles, Alexander~G Hauptmann, and Li
  Fei-Fei.
\newblock Peeking into the future: Predicting future person activities and
  locations in videos.
\newblock In {\em Proceedings of IEEE Conference on Computer Vision and Pattern
  Recognition (CVPR)}, 2019.

\bibitem{liu2020snce}
Yuejiang Liu, Qi Yan, and Alexandre Alahi.
\newblock Social nce: Contrastive learning of socially-aware motion
  representations.
\newblock In {\em Proceedings of International Conference on Computer Vision
  (ICCV)}, 2021.

\bibitem{loshchilov2018decoupled}
Ilya Loshchilov and Frank Hutter.
\newblock Decoupled weight decay regularization.
\newblock In {\em International Conference on Learning Representations (ICLR)},
  2018.

\bibitem{mangalam2020pecnet}
Karttikeya Mangalam, Harshayu Girase, Shreyas Agarwal, Kuan-Hui Lee, Ehsan
  Adeli, Jitendra Malik, and Adrien Gaidon.
\newblock It is not the journey but the destination: Endpoint conditioned
  trajectory prediction.
\newblock In {\em Proceedings of European Conference on Computer Vision
  (ECCV)}, 2020.

\bibitem{Marchetti_2020_CVPR}
Francesco Marchetti, Federico Becattini, Lorenzo Seidenari, and Alberto~Del
  Bimbo.
\newblock Mantra: Memory augmented networks for multiple trajectory prediction.
\newblock In {\em Proceedings of IEEE Conference on Computer Vision and Pattern
  Recognition (CVPR)}, 2020.

\bibitem{JSSv005i08}
George Marsaglia and Wai~Wan Tsang.
\newblock The ziggurat method for generating random variables.
\newblock {\em Journal of Statistical Software}, 2000.

\bibitem{samplingforqualitative}
Martin~N Marshall.
\newblock Sampling for qualitative research.
\newblock {\em Family Practice}, 1996.

\bibitem{5206641}
Ramin Mehran, Alexis Oyama, and Mubarak Shah.
\newblock Abnormal crowd behavior detection using social force model.
\newblock In {\em Proceedings of IEEE Conference on Computer Vision and Pattern
  Recognition (CVPR)}, 2009.

\bibitem{mohamed2020social}
Abduallah Mohamed, Kun Qian, Mohamed Elhoseiny, and Christian Claudel.
\newblock Social-stgcnn: A social spatio-temporal graph convolutional neural
  network for human trajectory prediction.
\newblock In {\em Proceedings of IEEE Conference on Computer Vision and Pattern
  Recognition (CVPR)}, 2020.

\bibitem{low_discrepancy}
Harald Niederreiter.
\newblock {\em Random Number Generation and Quasi-Monte Carlo Methods}.
\newblock Society for Industrial and Applied Mathematics, USA, 1992.

\bibitem{QMC}
Art~B. Owen.
\newblock Quasi-monte carlo sampling.
\newblock {\em Elsevier}, 2003.

\bibitem{QMC_faster_mc}
Spassimir~H. Paskov and Joseph~F. Traub.
\newblock Faster valuation of financial derivatives.
\newblock {\em The Journal of Portfolio Management}, 1995.

\bibitem{5459260}
Stefano Pellegrini, Andreas Ess, Konrad Schindler, and Luc Van~Gool.
\newblock You'll never walk alone: Modeling social behavior for multi-target
  tracking.
\newblock In {\em Proceedings of International Conference on Computer Vision
  (ICCV)}, 2009.

\bibitem{robicquet2016learning}
Alexandre Robicquet, Amir Sadeghian, Alexandre Alahi, and Silvio Savarese.
\newblock Learning social etiquette: Human trajectory understanding in crowded
  scenes.
\newblock In {\em Proceedings of European Conference on Computer Vision
  (ECCV)}, 2016.

\bibitem{rupprecht2017learning}
Christian Rupprecht, Iro Laina, Robert DiPietro, Maximilian Baust, Federico
  Tombari, Nassir Navab, and Gregory~D Hager.
\newblock Learning in an uncertain world: Representing ambiguity through
  multiple hypotheses.
\newblock In {\em Proceedings of International Conference on Computer Vision
  (ICCV)}, 2017.

\bibitem{sadeghian2019sophie}
Amir Sadeghian, Vineet Kosaraju, Ali Sadeghian, Noriaki Hirose, Hamid
  Rezatofighi, and Silvio Savarese.
\newblock Sophie: An attentive gan for predicting paths compliant to social and
  physical constraints.
\newblock In {\em Proceedings of IEEE Conference on Computer Vision and Pattern
  Recognition (CVPR)}, 2019.

\bibitem{salzmann2020trajectron++}
Tim Salzmann, Boris Ivanovic, Punarjay Chakravarty, and Marco Pavone.
\newblock Trajectron++: Dynamically-feasible trajectory forecasting with
  heterogeneous data.
\newblock In {\em Proceedings of European Conference on Computer Vision
  (ECCV)}, 2020.

\bibitem{shafiee2021Introvert}
Nasim Shafiee, Taskin Padir, and Ehsan Elhamifar.
\newblock Introvert: Human trajectory prediction via conditional 3d attention.
\newblock In {\em Proceedings of IEEE Conference on Computer Vision and Pattern
  Recognition (CVPR)}, 2021.

\bibitem{Shi2021sgcn}
Liushuai Shi, Le Wang, Chengjiang Long, Sanping Zhou, Mo Zhou, Zhenxing Niu,
  and Gang Hua.
\newblock Sgcn: Sparse graph convolution network for pedestrian trajectory
  prediction.
\newblock In {\em Proceedings of IEEE Conference on Computer Vision and Pattern
  Recognition (CVPR)}, 2021.

\bibitem{shi2020multimodal}
Xiaodan Shi, Xiaowei Shao, Zipei Fan, Renhe Jiang, Haoran Zhang, Zhiling Guo,
  Guangming Wu, Wei Yuan, and Ryosuke Shibasaki.
\newblock Multimodal interaction-aware trajectory prediction in crowded space.
\newblock In {\em Thirty-Fourth AAAI Conference on Artificial Intelligence},
  2020.

\bibitem{sun2020reciprocal}
Hao Sun, Zhiqun Zhao, and Zhihai He.
\newblock Reciprocal learning networks for human trajectory prediction.
\newblock In {\em Proceedings of IEEE Conference on Computer Vision and Pattern
  Recognition (CVPR)}, 2020.

\bibitem{sun2020rsbg}
Jianhua Sun, Qinhong Jiang, and Cewu Lu.
\newblock Recursive social behavior graph for trajectory prediction.
\newblock In {\em Proceedings of IEEE Conference on Computer Vision and Pattern
  Recognition (CVPR)}, 2020.

\bibitem{tao2020dynamic}
Chaofan Tao, Qinhong Jiang, and Lixin Duan.
\newblock Dynamic and static context-aware lstm for multi-agent motion
  prediction.
\newblock In {\em Proceedings of European Conference on Computer Vision
  (ECCV)}, 2020.

\bibitem{velivckovic2018graph}
Petar Veli{\v{c}}kovi{\'c}, Guillem Cucurull, Arantxa Casanova, Adriana Romero,
  Pietro Li{\`o}, and Yoshua Bengio.
\newblock Graph attention networks.
\newblock In {\em International Conference on Learning Representations (ICLR)},
  2018.

\bibitem{vemula2018social}
Anirudh Vemula, Katharina Muelling, and Jean Oh.
\newblock Social attention: Modeling attention in human crowds.
\newblock In {\em Proceedings of IEEE International Conference on Robotics and
  Automation (ICRA)}, 2018.

\bibitem{DC_MC_2}
Xin Xiong, Haipeng Xiong, Ke Xian, Chen Zhao, Zhiguo Cao, and Xin Li.
\newblock Sparse-to-dense depth completion revisited: Sampling strategy and
  graph construction.
\newblock In {\em Proceedings of European Conference on Computer Vision
  (ECCV)}, 2020.

\bibitem{MC_3Drecon}
Yifan Xu, Tianqi Fan, Yi Yuan, and Gurprit Singh.
\newblock Ladybird: Quasi-monte carlo sampling for deep implicit field based 3d
  reconstruction with symmetry.
\newblock In {\em Proceedings of European Conference on Computer Vision
  (ECCV)}, 2020.

\bibitem{yamaguchi2011you}
Kota Yamaguchi, Alexander~C Berg, Luis~E Ortiz, and Tamara~L Berg.
\newblock Who are you with and where are you going?
\newblock In {\em Proceedings of IEEE Conference on Computer Vision and Pattern
  Recognition (CVPR)}, 2011.

\bibitem{yi2015understanding}
Shuai Yi, Hongsheng Li, and Xiaogang Wang.
\newblock Understanding pedestrian behaviors from stationary crowd groups.
\newblock In {\em Proceedings of IEEE Conference on Computer Vision and Pattern
  Recognition (CVPR)}, 2015.

\bibitem{yu2020spatio}
Cunjun Yu, Xiao Ma, Jiawei Ren, Haiyu Zhao, and Shuai Yi.
\newblock Spatio-temporal graph transformer networks for pedestrian trajectory
  prediction.
\newblock In {\em Proceedings of European Conference on Computer Vision
  (ECCV)}, 2020.

\bibitem{yuan2021agent}
Ye Yuan, Xinshuo Weng, Yanglan Ou, and Kris Kitani.
\newblock Agentformer: Agent-aware transformers for socio-temporal multi-agent
  forecasting.
\newblock In {\em Proceedings of International Conference on Computer Vision
  (ICCV)}, 2021.

\bibitem{zhang2019sr}
Pu Zhang, Wanli Ouyang, Pengfei Zhang, Jianru Xue, and Nanning Zheng.
\newblock Sr-lstm: State refinement for lstm towards pedestrian trajectory
  prediction.
\newblock In {\em Proceedings of IEEE Conference on Computer Vision and Pattern
  Recognition (CVPR)}, 2019.

\bibitem{zhao2020tnt}
Hang Zhao, Jiyang Gao, Tian Lan, Chen Sun, Benjamin Sapp, Balakrishnan
  Varadarajan, Yue Shen, Yi Shen, Yuning Chai, Cordelia Schmid, Congcong Li,
  and Dragomir Anguelov.
\newblock Tnt: Target-driven trajectory prediction.
\newblock {\em In Conference on Robot Learning (CoRL)}, 2020.

\bibitem{zhao2019matf}
Tianyang Zhao, Yifei Xu, Mathew Monfort, Wongun Choi, Chris Baker, Yibiao Zhao,
  Yizhou Wang, and Ying~Nian Wu.
\newblock Multi-agent tensor fusion for contextual trajectory prediction.
\newblock In {\em Proceedings of IEEE Conference on Computer Vision and Pattern
  Recognition (CVPR)}, 2019.

\bibitem{MC_motionTracking}
Xiuzhuang Zhou and Yao Lu.
\newblock Abrupt motion tracking via adaptive stochastic approximation monte
  carlo sampling.
\newblock In {\em Proceedings of IEEE Conference on Computer Vision and Pattern
  Recognition (CVPR)}, 2010.

\bibitem{OKTEN20111268}
Giray Ökten and Ahmet Göncü.
\newblock Generating low-discrepancy sequences from the normal distribution:
  Box–muller or inverse transform?
\newblock {\em Mathematical and Computer Modelling}, 2011.

\end{thebibliography}
}

\end{document}